\documentclass[journal]{IEEEtran}

\usepackage{amsmath}
\newcommand\numberthis{\addtocounter{equation}{1}\tag{\theequation}}
\usepackage{algorithm}
\usepackage[noend]{algpseudocode}
\usepackage{amsthm}
\usepackage{glossaries}
\usepackage{amsfonts}

\newtheoremstyle{named}{}{}{\itshape}{}{\bfseries}{.}{.5em}{\thmnote{#3 }#1}
\theoremstyle{named}

\makeatletter
\def\BState{\State\hskip-\ALG@thistlm}
\makeatother
\usepackage{tabularx}
\usepackage{float}
\usepackage{graphicx}  % allows inclusion of PDF, PNG and JPG images
\usepackage[11pt]{moresize}
\ifCLASSINFOpdf
\else
\fi
\newcommand{\tablehere}{\begin{table}[H]}
\newcommand{\figurehere}{\begin{figure}[H]}
\newcommand{\graphicss}{\includegraphics[width=0.35\linewidth]} %0.45

\newcommand{\graphicssss}{\includegraphics[width=0.46\linewidth]} % 0.7
\newcommand{\graphicsssss}{\includegraphics[width=0.9\linewidth]} 

\begin{document}

\newacronym{sds}{SDS}{spoken dialogue system}
\newacronym{ai}{AI}{artificial intelligence}
\newacronym{rl}{RL}{reinforcement learning}
\newacronym{nn}{NN}{neural network}
\newacronym{pomdp}{POMDP}{partially observable Markov decision process}
\newacronym{mdp}{MDP}{Markov decision process}
\newacronym{nlg}{NLG}{natural language generation}
\newacronym{td}{TD}{temporal difference}
\newacronym{is}{IS}{importance sampling}
\newacronym{msve}{MSVE}{mean squared value error}
\newacronym{sgd}{SGD}{stochastic gradient descent}
\newacronym{nac}{NAC}{natural actor critic}
\newacronym{dqn}{DQN}{deep Q-network}
\newacronym{a2c}{A2C}{advantage actor critic}
\newacronym{fim}{FIM}{Fisher information matrix}
\newacronym{enac}{eNAC}{episodic natural actor critic}
\newacronym{ml}{ML}{machine learning}
\newacronym{gp}{GP}{Gaussian process}
\newacronym{drl}{DRL}{deep reinforcement learning}
\newacronym{er}{ER}{experience replay}
\newacronym{trpo}{TRPO}{trust region policy optimisation}
\newacronym{acer}{ACER}{actor critic with experience replay}
\newacronym{relu}{ReLU}{rectified linear unit}
\newacronym{kl}{KL}{Kullback-Leibler}
\newacronym{kkt}{KKT}{Karush-Kuhn-Tucker}
\title{Sample efficient deep reinforcement learning for dialogue systems with large action spaces}

\author{Gell\'{e}rt Weisz,
        Pawe\l~Budzianowski,~\IEEEmembership{Student Member,~IEEE,}\\
                Pei-Hao Su,~\IEEEmembership{Student Member,~IEEE,}
        and~Milica Ga\v{s}i\'{c},~\IEEEmembership{Member,~IEEE}
       
\thanks{This work was supported by EPSRC
Council and Toshiba Research Europe Ltd, Cambridge Research Laboratory.}% <-this % stops a space
\thanks{G. Weisz was with the  Department  of  Engineering,  University  of  Cambridge,  Trumpington  St,  Cambridge,  CB2  1PZ,  UK. He is now with DeepMind Technologies, London, UK, e-mail: gellert.weisz@gmail.com}
\thanks{P. Budzianowski, P.-H. Su. and M. Ga\v{s}i\'{c} are  with  the  Department  of  Engineering,  University  of  Cambridge,  Trumpington  St,  Cambridge,  CB2  1PZ,  UK,  e-mail:
\{pfb30,phs26,mg436\}
@cam.ac.uk}}
%\thanks{Manuscript received October 2017.}}


\markboth{}%
{Shell \MakeLowercase{\textit{et al.}}: Bare Demo of IEEEtran.cls for IEEE Journals}



\maketitle

\begin{abstract}
In spoken dialogue systems, we aim to deploy artificial intelligence to build automated dialogue agents that can converse with humans. A part of this effort is the policy optimisation task, which attempts to find a policy describing how to respond to humans, in the form of a function taking the current state of the dialogue and returning the response of the system. In this paper, we investigate deep reinforcement learning approaches to solve this problem. Particular attention is given to actor-critic methods, off-policy reinforcement learning with experience replay, and various methods aimed at reducing the bias and variance of estimators. When combined, these methods result in the previously proposed ACER algorithm that gave competitive results in gaming environments. These environments however are fully observable and have a relatively small action set so in this paper we examine the application of ACER to dialogue policy optimisation. We show that this method beats the current state-of-the-art in deep learning approaches for spoken dialogue systems. This not only leads to a more sample efficient algorithm that can train faster, but also allows us to apply the algorithm in more difficult environments than before. We thus experiment with learning in a very large action space, which has two orders of magnitude more actions than previously considered. We find that ACER trains significantly faster than the current state-of-the-art.\end{abstract}


% \fi
\IEEEpeerreviewmaketitle

\section{Introduction}
\IEEEPARstart{T}{raditionally}, computers are operated by either a keyboard and a mouse or touch. They provide feedback to the user primarily via visual clues on a display. This human-computer interaction model can be unintuitive to a human user at first, but it allows the user to express its intent clearly, as long as their goal is supported and they are equipped with sufficient knowledge to operate the machine. A \acrfull{sds} aims to make the human-computer interaction more intuitive by equipping computers with the ability to translate between human and computer language, thereby relieving humans of this burden and creating an intuitive interaction model. More specifically, the objective of an \acrshort{sds} is to help a human user achieve their goal in a specific domain (eg. hotel booking), using speech as the form of communication. Recent advances in \acrfull{ai} and \acrfull{rl} have established the necessary technology to build the first generation of commercial \acrlong{sds}s deployable as regular household items. Examples of such systems are Amazon's Alexa, Google's Home or Apple's Siri. While initially built as voice-command systems, over the years these systems have become capable of sustaining dialogues that can span a few turns.

Spoken dialogue systems are complex as they have to solve many challenging problems at once, under significant uncertainty. They have to recognise spoken language, decode the meaning of natural language, understand the user's goal while keeping track of the history of a conversation, determine what information to convey to the user, convert that information into natural language, and synthesise the sentences into speech that sounds natural. This work focuses on one particular step in this pipeline: devising a policy that determines the information to convey to the user, given our belief of their goal. 

This policy has been traditionally planned out by hand using flow-charts. This was a manual and inflexible process with many drawbacks that ultimately led to systems that were unable to converse intelligently.
To overcome this, the policy optimisation problem has been formulated as a reinforcement learning problem~\cite{roy2000spoken,young2002talking,young2013pomdp}. In this formulation, the computer takes actions and gets rewards. An algorithm aims to learn a policy that maximises the rewards through learning to take the best actions based on the state of the dialogue. Since the number of possible states can be very large (potentially infinite), complex and universal function approximators such as \acrlong{nn}s have been deployed as the policy~\cite{williams2017hybrid,su2017sample,fatemi2016policy}. There is a recent trend in the last years to model text-to-text dialogues with a neural network and tackle it as a sequence to sequence model. Initial attempts to do this underestimate the fact that planning is needed and treat the problem in a purely supervised fashion~\cite{wvmg17}. More recently RL learning has also been applied yielding improvements~\cite{zhes16,dllg17,lclg17,lila17}. While we focus here on traditional modular approaches, everything that we describe is also applicable to end-to-end modelling.

Using \acrlong{nn}s for policy optimisation is challenging for two reasons. First, there is often little training data available for an \acrshort{sds} as the data often comes from real humans. The system should be able to train quickly in an on-line setting while the training data is being gathered from users, to make the data to be gathered useful. Neural networks often exhibit too much bias or high variance when the volume of training data is small, making it difficult to quickly train them in a stable way. 
Second, the success or failure of a dialogue may be the only information available to the system to train the policy on. Dialogue success depends crucially on most actions in the dialogue, making it difficult to determine which individual actions contributed to the success, or led to the failure of a dialogue. This problem is exacerbated by the large size of the \emph{state space}: the system will potentially never be in the same state twice.

We address the above problems in the following ways:
\begin{enumerate}
\item We deploy the \acrshort{acer} algorithm~\cite{main,retrace} which has previously showed promising results on simpler gaming tasks.
\item We analyse the algorithm detail highlighting its theoretical advantages: low variance, safe and efficient learning.
\item We test the algorithm on a dialogue task with delayed rewards and test it alongside state-of-the-art methods in this task
\item After confirming its supremacy on a small action space, we deploy the algorithm on a two orders of magnitude larger action space.
\item We confirm our findings in a human evaluation.
\end{enumerate}
%This paper deploys \acrlong{nn}s for policy optimisation, aiming to solve the related problems and derive a quick and stable learning algorithm, relying on recent innovations in the field. The deployed algorithm, called \acrshort{acer}~\cite{main,retrace}, achieves the best results seen so far on \acrlong{nn}-based \acrlong{sds}s in the PyDial framework~\cite{ultes2017pydial}. We also investigate how both the state-of-the-art algorithm GP-SARSA~\cite{gasic2014gaussian} and \acrshort{acer}, can be optimised for a very large action space. To the best of our knowledge, this is the first time an RL algorithm is able to learn from scratch a policy for a large action set in a dialogue management task.

The rest of the paper is organised as follows. First, we give a brief introduction to dialogue management and define the main concepts. Then, in section~\ref{rl-intro} we review reinforcement learning. This is followed with an in-depth description of the ACER algorithm in section~\ref{acer}. Then, in section~\ref{acer-dm} we describe the architecture deployed to allow the application of ACER to a dialogue problem. The results of the extensive evaluation is given in section~\ref{eval}. In section~\ref{concl} we give conclusions and future work directions.

\section{Dialogue management}~\label{dm}

The job of the dialogue manager is to take the user's \emph{dialogue acts}, a semantic representation of the input, and determine the appropriate response also in the format of a \emph{dialogue act}~\cite{bunt94,trau00}. The function that chooses the appropriate response is called the \emph{policy}. The role of dialogue management is two-fold: tracking the dialogue  state and optimising the policy.

\subsection{Belief tracking}

We call the user's overall goal for a dialogue the \emph{user goal}, i.e.\ booking a particular flight or finding information about a restaurant. The user works towards this goal in every dialogue turn. In each dialogue turn, the short-term goal of the user is called the \emph{user intent}. Examples of \emph{user intent} are: \emph{confirm} what the system said, \emph{inform} the system on some criteria, and \emph{request} more information on something.

The belief tracker is the memory unit of the \acrshort{sds}, with the aim to track the \emph{user goal}, the \emph{user intent} and the \emph{dialogue history}. For the state to satisfy the Markov property it can only depend on the previous state and the action taken. Therefore, the state needs to encode enough information about what happened in the dialogue previously to maintain the conversation. By tracking the \emph{dialogue history} we ensure that the state satisfies the Markov property. The \emph{user intent} is derived from the (noisy) \emph{dialogue act}. To deal with the inherent uncertainty of the input, the dialogue is modelled as a \acrfull{pomdp}~\cite{young2013pomdp}. The \emph{belief state} is a vector representing a probability distribution over the different goals, intents and histories that occur in the dialogue. The role of the belief tracker is to accurately estimate this probability distribution and this is normally done using a version of a recurrent neural neural network~\cite{henderson2013deep}.

\subsection{Policy optimisation}
% if we have MDP, then state discrete and use tabular approach
% if we have a pomdp then b is continuous and we need to either discretise b and apply tabular approach or use fn approximation - could be linear (most common) or nonlinear nonparametric (GP), or NN
% credit assignment problem

A \emph{policy} is a probability distribution over possible \emph{user actions} given the current \emph{belief state}, and is commonly written as  $P_\pi(a|b) = \pi(a | b)$. Here, $a$ is the action and $b$ is the output of the belief tracker, which is interpreted as a vector of probabilities\footnote{Normally, in dialogue management, the action space is discrete.}.

In order to define the optimal policy, we need to introduce a utility function (\emph{reward}) that describes how good taking action $a$ is in state $b$. The reward for a complete dialogue depends on whether the user was successful in reaching their goal and the length of the conversation, such that short successful dialogues are preferred. Thus, the last dialogue interaction gains a reward based on whether the dialogue was successful, and every other interaction loses a small constant reward, penalising for the length of the dialogue. The task of policy optimisation is to maximise the expected cumulative reward in any state $b$ when following policy $\pi$, by choosing the optimal action $a$ from the set of possible actions $\mathbb{A}$. Finding the optimal policy is computationally prohibitive even for very simple POMDPs. We can view the \acrfull{pomdp} as a continuous-space \acrfull{mdp} in terms of policy optimisation, where the states are the \emph{belief states} \cite{kaelbling1998planning}. This allows us to apply function approximation to solve the problem of policy optimisation. This is the approach we adopt in this work.

\subsection{Action spaces}
\label{master-actions-section}

\emph{System actions} are the \emph{dialogue acts} that the system can give as a response. This is called the \emph{action space} or the \emph{master action space}. Due to its large size, training a dialogue policy in this action space is difficult. Some algorithms do not converge to the optimal policy, converge very slowly, or, in rare cases, have prohibitive computational demands\footnote{Since the training has to be on-line, i.e.\ happening while user input is acquired, training is constrained in computation time to prevent the user from having to wait for the system to reply. However, the training step is rarely the bottleneck.}. % question talk about GP master cubic cost >500ms

To alleviate this problem, we use the \emph{summary action space} which contains a much smaller number of actions. If a policy is trained on the \emph{summary action space}, the action selected by the policy needs to be converted to a \emph{master action}. The conversion is a set of heuristics that attempts to find the optimal slots to inform on given the belief state. 

Using the \emph{summary action space} provides  the clear benefit of a simpler dialogue policy optimisation task. On the other hand, the necessary heuristics to map to the \emph{master action space} need to be manually constructed for each domain. This means that the belief state needs to be human interpretable. This limits the applicability of neural networks for \emph{belief tracking} where the belief state is compactly represented as a hidden layer in the neural network.

The description of the summary and master actions that we consider is given in Appendix~\ref{caminfo}.

% talk about summary and master action space, pros and cons, and mapping between them, including a diagram

\subsection{Execution mask} \label{execmaskintro}
Not every system action is appropriate in every situation (belief state). For example, \emph{inform} is not a valid action at the very beginning of the dialogue, when the system has not yet received any information on what kind of entity the user is looking for. 
%Thus, we can simplify the action space further by restricting the system actions that could be selected to the actions that the system is able to produce a corresponding sensible response for. 
An \emph{execution mask} is constructed by the designer that ensures that only valid actions are selected by the policy: the probability of invalid actions is set to zero. The execution mask depends on the current belief state. Note that removing this mask inherently complicates the task of policy learning, as the policy then has to learn not to select  inappropriate actions based on the belief state.

\section{Reinforcement learning}\label{rl-intro}

In \acrlong{rl}, an agent interacts with the environment in discrete time steps. In each time step, the agent observes the environment as a belief state vector $b_t$ and chooses an action $a_t$ from the action space $\mathbb{A}$.  After performing action $a_t$, the agent observes a reward $r_t$ produced by the environment. 

The cumulative discounted return is the future value of an episode of interactions. For the $t^{\text{th}}$ timestep, we calculate this as
$$R_t = \sum_{i\geq 0} \gamma^i r_{t+i}.$$
The \emph{discount factor} $\gamma$ trades-off the importance of immediate and future rewards.
The goal of the agent is to find a policy that maximises the expected discounted cumulative return for every state. We define the value of a state-action pair $(b_t, a_t)$ under policy $\pi$ to be the Q-value function, the expectation of the return for belief state $b_t$ and action $a_t$:
$$
Q_\pi(b_t, a_t) = \mathop{\mathbb{E}_{\pi}}\big(R_t | b_t, a_t\big),
$$ %_{b_{t+1}:T, a_{t+1}:T
and the value of a state $b_t$ is the value function, which is the expected return only conditioned on the belief state $b_t$:
$$
V_\pi(b_t) = \mathop{\mathbb{E}}_{\pi}\big(R_t | b_t\big).
$$
In both definitions, the expectation is taken over the states the environment could be in after performing the current action, and the future actions selected by policy $\pi$ \cite{sutton1998reinforcement}.

As the reinforcement learning scenarios become more challenging,  the agent estimates value functions from trial and error by interacting with a simulated or real environment. These estimates are accurate only in the limit of infinite observations for each state-action pair, thus a requirement for the \emph{behaviour policy} $\mu$ is to maintain exploration, i.e.\ to keep visiting all state-action pairs with non-zero probability. The \emph{behaviour policy} is the policy used to generate the data during learning. For \textbf{on-policy methods}, the behaviour policy is the same as the learned policy $\pi$, in other words, we evaluate and improve the same policy that is used to make decisions. In contrast, \textbf{off-policy methods} evaluate and improve a policy different from the one used to generate the data, i.e.\ the behaviour policy and the learned policy can be different. The advantage of off-policy methods is that the optimal policy can be even while we are choosing sub-optimal actions according to the behaviour policy. 

Standard reinforcement learning algorithms require that the state space is discrete. Therefore, the belief state of a dialogue manager is often discretised to allow standard algorithms to be applied~\cite{gkms08}. Alternatively, function approximation can be applied for $Q$, $V$ or $\pi$.
In~\cite{chgp10}, linear function approximation was applied to value functions. As parametric function approximation can limit the optimality of the solution, GP-SARSA~\cite{gasic2014gaussian} instead models the Q-function as a Gaussian process. The key here is the kernel function which models the correlation between different states and allows uncertainty to be estimated. This is crucial for learning from a small number of samples. More recently, neural network function approximation was used to approximate the Q-value function, known as deep Q-learning~(DQN), obtaining human-level performance across challenging video games \cite{atari-deeprl-deepQnetwork}.
The policy $\pi$ can be also modelled directly by deep networks leading to the resurgence of actor-critic methods \cite{trpo}. The actor is improving its policy $\pi$ through interactions being directed by the critic (value function $V$).

\subsection{Neural networks in dialogue management}
A number of deep learning algorithms were previously applied to dialogue management. It has been shown in \cite{cuayahuitl2015strategic} that DQN enables learning strategic agents with negotiation abilities operating on a high-dimensional state space. 
 The performance of actor-critic models on task-oriented dialogue systems was analysed in \cite{fatemi2016policy}. These models can also be naturally bootstrapped with a small number of dialogues via supervised pre-training. They reported superior performance compared to GP-SARSA in a noise-free environment. However, the compared GP-SARSA did not utilise uncertainty estimates which were previously found to be crucial for effective learning~\cite{gasic2014gaussian}. 

Uncertainty estimates can be incorporated into DQN using Bayes-by-Backprop \cite{lipton2016efficient}. Initial results show an improvement in learning efficiency compared to vanilla DQN. %However, the overall performance in a dialogue system was very low.

A number of recent works investigated end-to-end using gradient descent techniques
 \cite{williams2016end,asadi2016sample,wvmg17}, where belief tracking and policy optimisation are optimised jointly. While, end-to-end modelling goes beyond the scope of this work, we note that the presented algorithm is applicable also in that setting.

\section{ACER}\label{acer}
This paper builds on recent breakthroughs in \acrfull{drl} and applies them to the problem of dialogue management. In particular, we investigate recently proposed improvements to the actor-critic method~\cite{main}. The goal is a stable and sample efficient learning algorithm that performs well on challenging policy optimisation tasks in the \acrshort{sds} domain. Recent advances in \acrshort{drl} apply several methods, including experience replay \cite{lin1992self}, truncated importance sampling with bias correction \cite{main}, the off-policy Retrace algorithm \cite{retrace} and trust region policy optimisation \cite{trpo} to various challenging problems. The core of this paper is to investigate to what extent these advances are applicable to the dialogue policy optimisation task with a large action space. These methods were recently combined in the \acrfull{acer} algorithm and tested in gaming environments. To this end, we explain \acrfull{acer} in detail and investigate the steps needed to apply it to SDS. Unlike in games, where these methods have been previously applied, we investigate dialogues with large and uncertain belief states and very large action spaces. This necessitates function approximation in reinforcement learning, but previously examined methods in \acrshort{sds} are data-inefficient, unstable or computationally too expensive. We investigate \acrshort{acer} as a means to overcome these limitations.
% We thus achieve stable and efficient learning, which ultimately allows larger action spaces to be considered. ACER proves to be surprisingly effective, beating the state of the art neural Network-based dialogue optimiser \cite{su2017sample}. This success motivated the extension that considers the more challenging domain of the \emph{master action space}. We conclude with a description of modifications to ACER and GP that have been devised to optimise the algorithms for master action space.
%As an extension to the project, 
% summarise main achievements

\subsection{Actor-critic with Experience Replay}
%\acrshort{dqn} samples its experience from a memory (\acrlong{er}), thus overcoming the correlated states and targets problem. However, it only estimates the $Q$-value function, leading to unstable learning. On the other hand, \acrshort{a2c} is an actor-critic method and it estimates both the \emph{value function} and the \emph{policy} \cite{mnih2016asynchronous}. Its targets are calculated from an unbiased, low-variance estimate, leading to more stable learning. \acrshort{a2c} does not have \acrlong{er} however, which means that only one update step can be made per iteration, leading to slower learning. To add \acrlong{er} to \acrshort{a2c}, an off-policy version of \acrshort{a2c} was derived~\cite{main}.

% introduce naive way of doing it
%To derive this mathematically, we consider the result from the Policy Gradient Theorem, which calculates the gradient as:
%$$\nabla J(\omega) = \sum_b d_\pi(b) \sum_a Q_{\pi}(b, a) \nabla_\omega \pi(a| b, \omega),$$
%where for a policy $\pi$, $$d^\pi(b) = \lim_{t\to\infty} P(b_t=b | b_0, \pi).$$
%We are sampling from the replay memory, and the policy $\mu$ (rather than $\pi$) was used to gather the episode, so we need to change the state weights to $d^\mu$ in the gradient equation: \cite{off-policy-ac}
%$$\nabla J(\omega) = \sum_b d_\mu(b) \sum_a Q_{\pi}(b, a) \nabla_\omega \pi(a| b, \omega).$$
In order to use experience replay in an actor-critic method, an off-policy version of the actor-critic method is needed. The objective is to find a policy that maximises the expected discounted return. This is equivalent to maximising the value of the initial state with input parameter vector $\omega$: $$J(\omega) = V_{\pi(\omega)}(b_0).$$ 
Another way of expressing the same objective is to maximise the cumulative reward received from the average state \cite{off-policy-ac}.
For behaviour policy $\mu$, let the occupancy frequency $d^\mu$ be defined as:
$$d^\mu(b) = \lim_{t\to\infty} P(b_t=b | b_0, \mu).$$
According to the new definition of $J(\omega)$, $V$ is weighted by $d^\mu$ because $\mu$ was used to collect the experience:
$$J(\omega) = \sum_{b\in\mathbb{B}} d^\mu(b) V_{\pi(\omega)} (b),$$
where $\pi$ is the optimal policy.

The off-policy version of the Policy Gradient Theorem \cite{sutton1999policy} is used to derive the gradients $\nabla_\omega J(\omega) \approx g(\omega) $:
%$$\begin{aligned}
%&\nabla_\omega J(\omega) \\
%&= \nabla_\omega \left[ \sum_{b\in\mathbb{B}} d^\mu(b) \sum_{a\in\mathbb{A}} \pi(a | b) Q_\pi (b, a)\right]\\
%&=\sum_{b\in\mathbb{B}} d^\mu(b) \sum_{a\in\mathbb{A}}\Big[ \nabla_\omega \pi(a|b) Q_\pi(b, a) + \pi(a | b) \nabla_\omega Q_\pi(b, a)  \Big].
%\end{aligned}$$
%The second term can be omitted \cite{off-policy-ac}. Thus, we %arrive at:
\begin{equation}\label{offpolicy-gradient}
g(\omega) = \sum_{b\in\mathbb{B}} d^\mu(b) \sum_{a\in\mathbb{A}} \nabla_\omega \pi(a|b) Q_\pi(b, a)
\end{equation}

%where
%\begin{equation}\label{offpolicy-gradient}
%g(\omega) = \sum_{b\in\mathbb{B}} d^\mu(b) \sum_{a\in\mathbb{A}} \nabla_\omega \pi(a|b) Q_\pi(b, a)
%\end{equation}

The states are encountered in proportions according to $d^\mu$ just by sampling from the experience memory, so there is no need to estimate $d^\mu$ explicitly. Estimating $Q_\pi$, however, is more difficult: the off-policy interactions are gathered according to $\mu$, and we need the $Q$-function under a different - current policy $\pi$. %We will look at different methods to estimate $Q_\pi$.

%Na\"ively, we can estimate $Q_{\pi}(b, a)$ with an importance sampled version of the discounted cumulative reward $R$, sampled from the replay memory:
%$$\nabla J(\omega) \approx \mathop{\mathbb{E}}\left[\left(\prod_{t=0}^T \rho_t\right)\left(\sum_{i=0}^T \gamma^i r_{t+i}\right) \nabla_\omega \pi(a| b, \omega) | b\sim d^\mu\right].$$
%where $\rho_t = \frac{\pi(a_t | b_t)}{\mu(a_t | b_t)}$ are the \acrfull{is} weights \cite{precup2001off}. This estimation is unbiased, but suffers from very high variance \cite{main}. This is because it multiplies potentially unbounded \acrshort{is} weights for an entire episode. This multiplication either results in a very small value (\emph{vanishing weight}) or a very large value (\emph{exploding weight}). 
To account for this, the \acrfull{is} weights \cite{precup2001off} could be used. To achieve stable learning, we use an estimation method that achieves low variance by considering state-action pairs in isolation, applying only one \acrshort{is} weight for each.% thus avoiding the high variance commonly observed for estimators that ...

%Starting again from the off-policy version of the policy gradient theorem:
Continuing from Equation~(\ref{offpolicy-gradient}), the approximation of the true gradient can be derived:
$$\begin{aligned}
g(\omega) %&= \sum_{b\in\mathbb{B}} d^\mu(b) \sum_{a\in\mathbb{A}} \nabla_\omega \pi(a|b) Q_\pi(b, a)\\
&= \mathop{\mathbb{E}} \left[  \sum_{a\in\mathbb{A}} \nabla_\omega \pi(a|b) Q_\pi(b, a) | b\sim d^\mu \right]\\
&= \mathop{\mathbb{E}} \left[  \sum_{a\in\mathbb{A}} \mu(a | b) \frac{\pi(a | b)}{\mu(a | b)}\frac{\nabla_\omega \pi(a|b)}{\pi(a | b)} Q_\pi(b, a) | b\sim d^\mu \right]\\
&= \mathop{\mathbb{E}} \left[ \rho(a | b) \nabla_\omega \log \pi(a|b) Q_\pi(b, a) | b\sim d^\mu, a\sim \mu(\cdot|b) \right],\\
\end{aligned}$$
where $\rho(a|b) = \frac{\pi(a | b)}{\mu(a | b)}$ are the \acrfull{is} weights.
The advantage function $A_\theta$ is used in place of the $Q$-function for an unbiased estimate with a lower variance:
$$g(\omega) = \mathop{\mathbb{E}}_\mu \left[ \rho(a | b) \nabla_\omega \log \pi(a|b) A_\theta\right].$$
%The advantage function is approximated in the vanilla version of \acrshort{a2c} as $R_t - V(b_t, \theta)$. We cannot use this here as the cumulative reward $R$ has been gathered according to the old policy $\mu$ and may not be representative of the current cumulative reward we obtain following $\pi$. 
In off-policy setting, the advantage function is approximated as as $r_t + \gamma V(b_{t+1}, \theta) - V(b_t, \theta)$.% This also applies to the loss of the critic, which used to be $\left(R_t - V(b_t, \theta)\right)^2$. \acrshort{a2c} with \acrshort{er} uses $\left(r_t + \gamma V(b_{t+1}, \theta) - V(b_t, \theta)\right)^2$ instead. %Algorithm~\ref{a2c-er} shows the complete algorithm.

% \begin{algorithm}
% \caption{\acrlong{a2c} with \acrlong{er}}\label{a2c-er}
% \begin{algorithmic}[1]
% \item Input: policy $\pi(a|b, \omega)$, $V_\theta(b, a)$, learning rate $\alpha$
% \item Initialise $\theta, \omega, V_\theta(\mbox{terminal}) = 0$
% \Repeat
% \State Generate an episode according to $\pi(\cdot|\cdot, \omega)$ and save to replay memory
% \For{$i=0, 1, ..., \mbox{batch\_size}$}
% \State Sample episode $\{b_{0:T}, a_{0:T}, r_{0:T}\}$ from replay memory, with old policy $\mu(a | b)$
% \For{$t=T\mbox{ downto } 0$}
% \State $R_{t-1} \gets r_t + \gamma V(b_t, \theta)$
% \State $\rho(a_t| b_t) \gets \frac{\pi(a_t | b_t)}{\mu(a_t | b_t)}$
% \State $\nabla J = \nabla J + \rho(a_t | b_t)\left(r_t + \gamma V(b_{t+1}, \theta) - V(b_t, \theta)\right)\nabla_\omega \log \pi(a_t | b_t, \omega)$
% \State $\nabla L = \nabla L + \nabla_\theta\left(r_t + \gamma V(b_{t+1}, \theta) - V(b_t, \theta)\right)^2$
% \EndFor
% \EndFor
% \State $\omega \gets \omega + \alpha \nabla J$
% \State $\theta \gets \theta + \alpha \nabla L$
% \Until convergence
% \end{algorithmic}
% \end{algorithm}

\subsection{Lambda returns}
The unbiased estimator $$Q_\pi(b_t, a_t)\approx \sum_{i=t}^T \gamma^i r_i$$ results in high variance, due to the \acrshort{is} weight that has to be calculated for the entire episode. The \acrfull{td} estimation 
\begin{equation}\label{eq:q_est}
Q_\pi(b_t, a_t)\approx r_t + \gamma V(b_{t+1}, \theta)
\end{equation} 
only requires a single \acrshort{is} weight. However, this estimation is biased: 
the value function update of the current state is based on the current estimate of the value function for the next state. This leads to slow convergence or no convergence at all.

It is possible to combine both methods and create an estimator that trades off bias and variance according to a parameter $\lambda$. \cite{off-policy-ac} estimate $Q_\pi$ as:
$$Q_\pi(b_t, a_t)\approx R_{t,\lambda}\mbox{, where}$$
$$\begin{aligned}
R_{T, \lambda} &= r_T\\
R_{t,\lambda}&=r_t + (1-\lambda)\gamma V(b_{t+1}) + \lambda \gamma \rho_{t+1} R_{t+1,\lambda}.
\end{aligned}$$
The constant $\lambda$ controls the bias-variance trade-off: setting $\lambda$ to 0 results in an equivalent estimation as in Equation~(\ref{eq:q_est}),
%what we had before: $Q_\pi(b_t, a_t)\approx r_t + \gamma V(b_{t+1}, \theta)$, 
with a low variance but high bias. Conversely, setting $\lambda$ to 1 results in high variance as many \acrshort{is} weights will be producted. This has the advantage of propagating the final reward further to the starting state which reduces bias. %A carefully hand-selected $\lambda$ could bring the best of both worlds.

%It is important to note that this approach has some shortcomings. First, it is required to set $\lambda$ ahead of time to represent a good trade-off. Second, even when $\lambda$ is 0 to reduce variance as much as possible, occasional large \acrshort{is} weights introduce the variance, and they can still cause instability \cite{main}.

\subsection{Retrace}

The Retrace algorithm (\cite{retrace}) attempts to estimate the current $Q$-function from off-policy interactions in a safe and efficient way, with small variance. Throughout this discussion, we call a method \emph{safe} if its estimate of $Q^\pi$ can be proven to converge to $Q^\pi$.
% what is safety?
The updated estimate of the $Q$-function, $Q^{ret}$ is computed based on state-action trajectories sampled from the replay memory:
\begin{align*} 
&Q^{ret}(b, a)= Q(b, a) + \\
&\mathop{\mathbb{E}}_\mu\left[ \sum_{t\geq 0} \gamma^t \left(\prod_{s=1}^t c_s\right) (r_t + \gamma V(b_{t+1}) - Q(b_t, a_t)) \right].\numberthis\label{retrace-eq}
\end{align*}
The methods that stem from this framework differ only in their definition of \emph{eligibility traces} $c_s$.

This framework introduces changes to the actor-critic model. Instead of approximating $V$ and $\pi$ with \acrshort{nn}s and estimating $Q$ in a closed-form equation to compute the update targets, both $\pi$ and $Q$ are estimated with \acrshort{nn}s. $V$ is then computed  from $\pi$ and $Q$:
\begin{equation}\label{value-calc}
V(b) = \mathop{\mathbb{E}}_\pi \left[ Q(b, \cdot) \right]= \sum_a \pi(a | b) Q(b, a).
\end{equation}

We focus on Retrace proposed by \cite{retrace} where $c_s = \lambda \min\left(1, \rho(a_s| b_s)\right)$.
Ideally, we need a method that is safe, has low variance and is as efficient as possible. Retrace solves this trade-off by setting the traces ``dynamically'', based on the \acrshort{is} weights. In the near on-policy case, it is efficient as \acrshort{is} weights will be about 1, preventing the traces from vanishing. It has low variance because the \acrshort{is} weights are clipped at 1. It is also safe for any $\pi$ and $\mu$. 
The goal of this discussion is limited to conveying the intuition behind \emph{Retrace}, but a full proof of safety is available in \cite{retrace}.

\subsection{Computational cost}
Let us investigate the computational cost of deriving $Q^{ret}$ from $Q$ in a na\"ive way. For each episode sampled from the replay memory, and for each state-action pair, we need to visit the remaining part of the episode to calculate the expectation of errors under $\mu$ according to Equation~(\ref{retrace-eq}). This quadratic element of the computational cost can be reduced to a linear one by deriving $Q^{ret}$ in a recursive way. For an episode trajectory $b_{1:T}, a_{1:T}$ sampled from the replay memory, Equation~(\ref{retrace-eq}) becomes:
$$\begin{aligned}
Q^{ret}(b_i, a_i) &= r_{i} + \gamma V(b_{i+1})\\
&\quad\;+\gamma c_{i+1}\left(Q^{ret}(b_{i+1}, a_{i+1}) - Q(b_{i+1}, a_{i+1})\right).\\
\end{aligned}$$
%$$\begin{aligned}
%&Q^{ret}(b_i, a_i) \\
%&=Q(b_i, a_i) + \sum_{t\geq 0}^{T-i} \gamma^t (\prod_{s=1}^t c_{i+s}) \\
%&\qquad\Big(r_{i+t} + \gamma V(b_{t+i+1}) - Q(b_{i+t}, a_{i+t})\Big)\\
%&= Q(b_i, a_i) + r_{i} + \gamma V(b_{i+1}) - Q(b_{i}, a_{i}) \\
%&\qquad+ \gamma c_{i+1} \sum_{t\geq 0}^{T-i-1}\gamma^t \left(\prod_{s=1}^t c_{i+1+s}\right)\\
%&\qquad\quad\Big(r_{i+1+t} + \gamma V(b_{t+i+2}) - Q(b_{i+1+t}, a_{i+1+t})\Big)\\
%&= r_{i} + \gamma V(b_{i+1})\\
%&\qquad+\gamma c_{i+1}\left(Q^{ret}(b_{i+1}, a_{i+1}) - Q(b_{i+1}, a_{i+1})\right).\\
%\end{aligned}$$
We will use this more computationally efficient, recursive formulation of $Q^{ret}$.

\subsection{Importance weight truncation with bias correction}

Currently, we calculate the policy gradient as:
\begin{equation}\label{what-we-had-before}
g(\omega) = \mathop{\mathbb{E}}_{b_t\sim d^\mu, a_t\sim \mu}\left[ \rho(a_t|b_t) \nabla_\omega \log \pi(a_t|b_t) A_\pi(b_t, a_t)\right],
\end{equation}
where the expectation is taken over the replay memory, and $\rho(a_t|b_t) = \frac{\pi(a_t|s_t)}{\mu(a_t|s_t)}$. An issue with this approximation is that the \acrshort{is} weights $\rho(a_t|b_t)$ are potentially unbounded, introducing significant variance. To solve this problem, we clip the \acrshort{is} weights from above by a constant $c$: $\overline{\rho}(a_t|b_t) = \min\{ c, \rho((a_t|b_t)) \}$. We can split the equation into two parts, one involving the truncated \acrshort{is} weight, and the other the residual. We also need to estimate the residual, otherwise we introduce bias in the gradient estimation. We call the residual the \emph{bias correction} term.
$$\begin{aligned}
g(\omega) &= %\approx %&= \mathop{\mathbb{E}}_{b_t\sim d^\mu, a_t\sim \mu}\left[ \rho(a_t|b_t) \nabla_\omega \log \pi(a_t|b_t) A_\pi(b_t, a_t)\right]\\
 \mathop{\mathbb{E}}_{b_t\sim d^\mu}\Big[ \mathop{\mathbb{E}}_{a_t\sim \mu}\overline{\rho}(a_t|b_t) \nabla_\omega \log \pi(a_t|b_t) A_\pi(b_t, a_t)\\
&\quad +  \mathop{\mathbb{E}}_{a_t\sim \mu}\left[\rho(a_t|b_t) -c \right]_+ \nabla_\omega \log \pi(a_t|b_t) A_\pi(b_t, a_t)\Big],
\end{aligned}$$
where $[\cdot]_+= \max(0, \cdot)$. The weight of the bias correction term, $\left[\rho(a_t|b_t) -c \right]_+$, can still be unboundedly large. This can be solved by sampling the action from the distribution $\pi$ rather than $\mu$ \cite{main}:
\begin{equation} \label{final}
\begin{split}
g(\omega) &=  \mathop{\mathbb{E}}_{b_t\sim d^\mu}\Bigg[ \mathop{\mathbb{E}}_{a_t\sim \mu}\overline{\rho}(a_t|b_t) \nabla_\omega \log \pi(a_t|b_t) A_\pi(b_t, a_t)\\
&\quad\quad +   \mathop{\mathbb{E}}_{a\sim \pi}\left[\frac{\rho(a|b_t) -c}{\rho(a|b_t)} \right]_+ \nabla_\omega \log \pi(a|b_t) A_\pi(b_t, a)\Bigg].
\end{split}
\end{equation}
There are two key advantages of this formulation: 
\begin{itemize}
\item The bias correction term ensures that the estimate of the gradient remains unbiased.
\item The bias correction term is only active when $\rho(a|b)>c$, and otherwise the formulation is equivalent to Equation~(\ref{what-we-had-before}). When active, the bias correction weight falls between 0 and 1.
\end{itemize}
%The bias correction term ensures that the estimate of the gradient remains unbiased. Furthermore, the bias correction term is only active when $\rho(a|b)>c$, and otherwise the formulation is equivalent to what we had before. 
%This means we can tune $c$ to a high enough value to only modify the handling of belief-action pairs with a high-variance \acrlong{is}. For these cases, when $\rho(a|b)>c$, the variance of the estimation is significantly reduced: the base term's weight is clipped by $c$, and the bias correction weight, $\left[\frac{\rho(a|b_t) -c}{\rho(a|b_t)} \right]_+$, falls between 0 and 1, both being bounded.

To apply this method, called the \emph{truncation with bias correction trick} by \cite{main}, we have to overcome a problem with the advantage function estimation. Before, we estimated $A_\pi(b, a) = Q^{ret}(b, a) - V(b) = Q^{ret}(b, a) - \sum_a \pi(a | b) Q(b, a)$ for belief-action pairs that we sampled from the replay memory, Equation~(\ref{value-calc}). 
For the bias correction term however, only the belief is sampled from the memory, and all the actions are considered and weighted by the current policy $\pi$. Due to the way $Q^{ret}$ is formulated, it learns from rewards, and only learns belief-action pairs that have been visited and sampled from the replay memory. Thus the estimation is not available for the bias correction term, so we use the output of the \acrshort{nn}, $Q$, to estimate the advantage function for that term: $A'_\pi(b, a) = Q(b, a) - \sum_a \pi(a | b) Q(b, a)$.

\subsection{Trust Region Policy Optimisation}
\label{section-beta}
\label{section-trpo}

% todo problem with paragraph: no citations but lots of statements, difficult to follow. with appropriate citations would be fine
Typically, the step size parameter in the gradient descent is calculated assuming the that the policy parameter space is Euclidian. However, this has a major shortcoming: small changes in the parameter space can lead to erratic changes in the output policy~\cite{thomson-thesis,pascanu2013revisiting}. This could lead to unstable learning or a learning rate too small for quick convergence. This is solved in the Natural Actor Critic algorithm by considering the \emph{natural gradient}~\cite{nac}. 
%However, the computation of the \emph{natural gradient} restricts the applicability of this algorithm.
%However, the training procedure for \acrshort{nac} is very different from the off-policy gradient estimation methods presented here. 
%Crucially, \acrshort{nac} is based on \emph{compatible function approximation} to avoid an expensive Fisher information matrix computation. This in turn requires a different architecture and training procedure from the off-policy gradient estimation methods presented here. Thus, computing the natural gradient for \acrshort{acer} efficiently would require significant further work. We could also attempt to modify the \acrshort{nac} algorithm (\cite{nac}) to work off-policy, but this also goes beyond the scope of this paper.

Instead of computing the exact natural gradient, we can approximate it. For the natural gradient, the distance metric tensor is the Fisher information matrix:
$$
(G_\omega)_{ij} = \mathop{\mathbb{E}}\left[ \frac{\delta \log p(x | \omega)}{\delta \omega_i} \frac{\delta \log p(x | \omega)}{\delta \omega_j}  \right].
$$
It can be shown \cite{kober2009policy} that
$$
d\omega^T G_\omega d \omega \approx
\mbox{KL}(\pi(\cdot | b, \omega) || \pi(\cdot| b, \omega + d \omega)),
$$
Where $\mbox{KL}$ is the Kullback-Leibler divergence. %It is defined for discrete probability distributions as
%$$\mbox{KL}(P||Q) = \sum_i P(i) \log\frac{P(i)}{Q(i)}.$$
Thus, instead of directly restricting the learning step-size with the \emph{natural gradient} method, we can approximate the same method by restricting the Kullback-Leibler divergence between the current policy $\pi$ parametrised by $\omega$, and the updated policy parametrised by $\omega + \alpha\cdot \nabla_\omega J$, for learning rate $\alpha$. %More specifically, we will update the parameters with a value as close as possible to $\omega + \alpha\cdot \nabla_\omega J$, such that the new policy is within a constant distance of the old policy, where the distance metric is \acrshort{kl} divergence. 
This method is called \emph{\acrfull{trpo}}, introduced by \cite{trpo}. Their method, however, relies on repeated computations of Fisher matrices for each update, which can be prohibitively expensive. \cite{main} introduces an efficient \acrshort{trpo} method that we will adopt instead. Our description of the method largely follows theirs with additional explanations and necessary adaptation to our discrete action-space \acrshort{sds} domain.

To begin with, \cite{main} proposes that the \acrshort{kl}-divergence to the updated policy should be measured not from the current policy, but from a separate average policy instead. This stabilises the algorithm by preventing it from gaining momentum in a specific direction. Instead, it is restricted to stay around a more stable average policy $\pi_a$. The average policy is parametrised with $\omega_a$, where $\omega_a$ represents a running average of all previous policy parameters. It is updated \emph{softly} after each learning step as:
$$\omega_a \gets \beta \omega_a + (1-\beta) \omega.$$
$\beta$ is a hyperparameter that controls the amount of history to maintain in the average policy. A value close to zero makes the average policy forget the history very quickly, reducing the effect of calculating the distances from the average policy instead of the current one. A value close to one will prevent the average policy to adjust to the current policy, or slows this adjustment process down. %However, the weight given to the history is reduced exponentially as the learning steps progress. A setting of $\beta$ between 0.95 and 0.99 turned out to work best in our case.
% \begin{figure}
% \centering
% \graphicsss{trpo3.png}
% \caption{
% \footnotesize
% Trust Region Policy Optimisation.}
% \label{fig:trpo}
% \end{figure}

% Next, we define our goal for \acrshort{trpo}. We work with our previous definition of the policy gradient:
% $$\begin{aligned}
% g(\omega) &= \mathop{\mathbb{E}}_{b_t\sim d^\mu}\Bigg[ \mathop{\mathbb{E}}_{a_t\sim \mu}\overline{\rho}(a_t|b_t) \nabla_\omega \log \pi(a_t|b_t)\\
% &\qquad\quad\left(Q^{ret}(b, a) - \sum_a \pi(a | b) Q(b, a)\right)\\
% &\qquad+   \mathop{\mathbb{E}}_{a\sim \pi}\left[\frac{\rho(a|b_t) -c}{\rho(a|b_t)} \right]_+ \nabla_\omega \log \pi(a|b_t) \\
% &\qquad\quad\left(Q(b, a) - \sum_a \pi(a | b) Q(b, a)\right)\Bigg].
% \end{aligned}$$
\acrshort{trpo} can be formulated as an optimisation problem, where we aim to find $z$ that minimises the L2-distance between $z$ and the vanilla gradient $g(\omega)$ from~\ref{final}. This is a quadratic minimisation. In addition, our aim is for the divergence constraint to be formulated in a linear way, which will allow to derive a closed-form solution. Since $z$ will be used for the parameter update, we have 
$\omega'=\omega + \alpha z$, where $\omega'$ denotes the updated parameter vector. We can approximate the \acrshort{kl} divergence after the policy update using a first-order Taylor expansion:
$$\begin{aligned}
&\mbox{KL}\left[ \pi(\cdot|\omega_a) || \pi(\cdot | \omega') \right] = \\
&\mbox{KL}\left[ \pi(\cdot|\omega_a) || \pi(\cdot | \omega) \right] + \nabla_\omega\mbox{KL}\left[ \pi(\cdot|\omega_a) || \pi(\cdot | \omega) \right]^T \cdot \alpha z.
\end{aligned}$$
So the increase in \acrshort{kl} divergence in this step is
$$\nabla_\omega\mbox{KL}\left[ \pi(\cdot|\omega_a) || \pi(\cdot | \omega) \right]^T \cdot \alpha z.$$
We can constrain this increase to be small by setting $\delta$, such that
$$\nabla_\omega\mbox{KL}\left[ \pi(\cdot|\omega_a) || \pi(\cdot | \omega) \right]^T \cdot z \leq\delta,$$
where the learning rate $\alpha$ is left out, since it is a constant and can be incorporated into $\delta$.
Letting $k=\nabla_\omega\mbox{KL}\left[ \pi(\cdot|\omega_a) || \pi(\cdot | \omega) \right]$, the optimisation problem with linearised \acrshort{kl} divergence constrain is \cite{main}:
\begin{equation*}
\begin{aligned}
& \underset{z}{\text{minimize}}
& & \frac{1}{2} || g(\omega) - z ||^2_2 \\
& \text{subject to}
& & k^T z \leq \delta
\end{aligned}
\end{equation*}
Since the constraint is linear, the overall optimisation problem reduces to a simple quadratic programming problem. Thus, a closed-form solution can be derived using the KKT conditions \cite{gellert}:
$$z = g - \max\{0 , \frac{k^T g - \delta}{||k||_2^2} k\}.$$

\subsection{Summary of \acrshort{acer}}

ACER is the result of all methods presented in this section. With on-policy exploration, it is a modified version of \acrshort{a2c}. Both ACER and A2C use \acrlong{er} and sample from their memories to achieve high sample efficiency. The difference between them is that \acrshort{acer} additionally employs \acrshort{trpo}, and that it uses a $Q$-function estimator instead of a $V$-function estimator as the critic. When off-policy, it uses truncated importance sampling with bias correction \cite{main} to reduce the variance of \acrshort{is} weights without adding bias. The Retrace algorithm is used to compute the targets based on the observed rewards in a safe, efficient way, with low bias and variance.  

This training algorithm is presented in pseudocode (Algorithm~\ref{acer-pseudo-2}), and is called from the master ACER algorithm (Algorithm~\ref{acer-pseudo}). It performs $\epsilon$-greedy exploration, i.e.\ the optimal action learned so far with probability $1-\epsilon$, and a random action with probability $\epsilon$. A hyperparameter $\mbox{batch\_size}$ controls the number of dialogues considered for a training step, and $n$ controls the number of training steps for each new dialogue gathered. We will investigate the effect of various hyperparameters and how to set them in Section~\ref{hyperparams}.

\begin{algorithm}
\caption{\acrshort{acer} master algorithm}\label{acer-pseudo}
\begin{algorithmic}[1]
\item Input: policy $\pi(a|b, \omega)$, $Q_\theta(b, a)$, hyperparameters $\alpha, \beta, \gamma, \delta$
\item Initialise $\theta, \omega, \omega_a$, and $Q_\theta(\mbox{terminal}) = 0$
\Repeat
\State Generate episode $\{b_{0:T}, a_{0:T}, r_{0:T}\}$ according to $\epsilon$-greedy using $\pi(\cdot|\cdot, \omega)$
\State Save generated episode, along with values of $\pi(\cdot|\cdot, \omega)$
\For{$i=1$ to $n$}
\State Sample a subset of replay memory, $M$, of size $\mbox{batch\_size}$
\State Call training algorithm (Algorithm~\ref{acer-pseudo-2}) with $\{M, \theta, \omega, \omega_a, \pi, Q, \alpha, \beta, \gamma, \delta\}$
\EndFor
\Until convergence
\end{algorithmic}
\end{algorithm}

\begin{algorithm}
\caption{\acrshort{acer} training algorithm}\label{acer-pseudo-2}
\begin{algorithmic}[1]
\item Input: $\{M, \theta, \omega, \omega_a, \pi, Q, \alpha, \beta, \gamma, \delta\}$
\item Initialise $g=0, \; d\theta=0$
\For{each dialogue $\{b_{1:N}, a_{1:N}, r_{1:N}, \mu\}$ in M}
	\For{$1=N$ to $1$}
		\State $\rho_t \gets \frac{\pi(a_t | b_t, \omega)}{\mu(a_t | b_t)}$
		\State $V(b_t) \gets \sum_a Q_\theta(b_t, a) \pi(a | b_t, \omega)$
		\State $Q^{ret} \gets r_t + \gamma Q^{ret}$

		\State $A_\pi(b_t, a_t) \gets Q^{ret} - V(b_t)$
		\State $A'_\pi(b_t, a_t) \gets Q_\theta(b_t, a_t) - V(b_t)$

		\State $\overline{\rho}_t \gets min(r, \rho_t)$
        \State $C\gets\left[\frac{\rho(a|b_t, \omega) -c}{\rho(a|b_t, \omega)} \right]_+$ 
		\State $B\gets \sum_{a}\pi(a | b_t, \omega)\: C\: \nabla_\omega \log \pi(a|b_t, \omega) A'_\pi(b_t, a)$
		\State $g \gets g + \overline{\rho}_t \nabla_\omega \log \pi(a_t|b_t, \omega) A_\pi(b_t, a_t)
		+ B$ %\Comment{IS truncation plus bias correction}

		\State $d\theta \gets d\theta - \nabla_\theta (Q^{ret} - Q_\theta(b_t, a_t))^2$
		\State $Q^{ret} \gets \overline{\rho}_t (Q^{ret} - Q_\theta(b_t, a) ) + V(b_t)$
	\EndFor
\EndFor
\State $k\gets\nabla_\omega\mbox{KL}\left[ \pi(\cdot|\omega_a) || \pi(\cdot | \omega) \right]$
\State $z\gets g - \max\{0 , \frac{k^T g - \delta}{||k||_2^2} k\}$
\State $\omega \gets \omega + \alpha\cdot z$
\State $\theta \gets \theta + \alpha\cdot d\theta$
\State $\omega_a \gets \beta \omega_a + (1-\beta) \omega$
\end{algorithmic}
\end{algorithm}

\section{ACER for dialogue modelling}\label{acer-dm}
\label{acer-arch-section}
% explain network and how to use retrace for updates
In this section we detail the steps needed to apply \acrshort{acer} to a dialogue task.
\subsection{Learning in summary action space}

\begin{figure}
\centering
	\graphicss{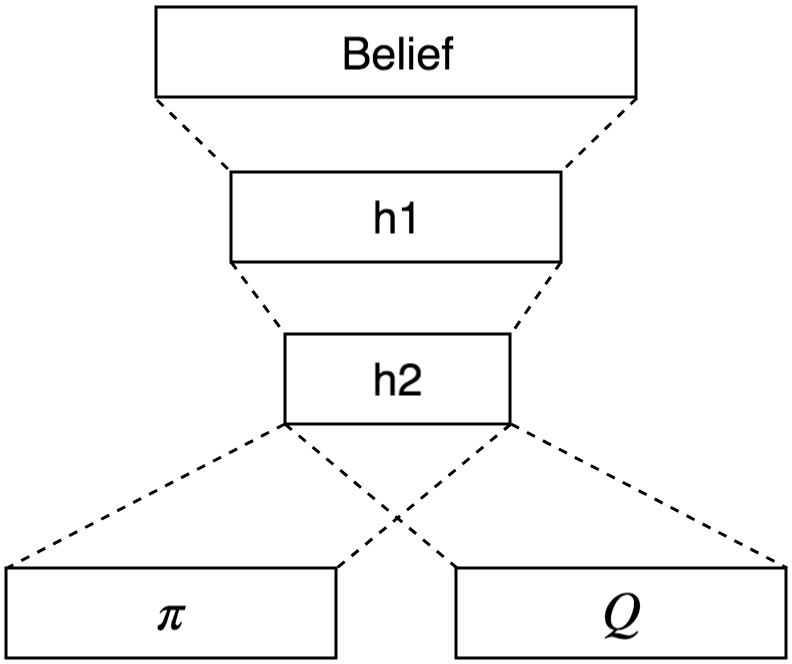}
\caption{ACER neural network architecture for dialogue management.}
% todo make smaller and maybe left-to-right
\label{fig:nn-arch}
\end{figure}

Let us design the \acrlong{nn}s for actor-critic for a dialogue management task. On top of the input of the belief state, we build two hidden layers, $h_1$ and $h_2$. The heads of the \acrshort{nn} are the functions $\pi$ and $Q$. 
Both hidden layers $h_1$ and $h_2$ are shared between the predictors of $Q$ and $\pi$. Weight sharing can be beneficial as it reduces the number of parameters to train. Furthermore, in a dialogue system, we expect a strong positive correlation between $\pi$ and $Q$. The architecture is illustrated in Figure~\ref{fig:nn-arch}.

%The layers are fully-connected between the input (belief) layer and $h_1$, as well as between $h_1$ and $h_2$. 
The activation function for layers $h_1$ and $h_2$ is \acrfull{relu}, which was chosen empirically as it led to faster training. The activation function for $\pi$ is \emph{softmax}, which converts the inputs to a probability distribution with values between 0 and 1, summing up to 1. There is no activation function for the output $Q$, as we want it to have an unlimited range, both from above and below (as rewards can be negative). All the connections in the \acrshort{nn} are fully connected, which imposes the least structural constraints on the architecture. We perform our experiments on the \emph{Cambridge Restaurants} domain, the details of which are given in Appendix~\ref{caminfo}. In this domain the \emph{belief state} is represented by a 268-dimensional vector.
%  You should explain (maybe via a diagram) what these 268 dimensions represent
This is the input of the \acrshort{nn}. Our layer $h_1$ consists of 130 neurons and $h_2$ has 50 neurons. These numbers were chosen empirically, with the goal in mind to force the \acrshort{nn} to encode all information about the \emph{belief state} relevant to $\pi$ and $Q$ in the bottleneck layer of 50 neurons, thereby learning a mapping that generalises better. The output vectors $\pi$ and $Q$ have the dimensionality of the action space. Initially, we experiment with the summary action space, which has 15 actions (see Appendix~\ref{caminfo} for details).

\subsection{Learning in master action space}

\subsubsection{Master actions for \acrshort{acer}}
\label{section-macer}

In addition to applying ACER on the summary space, we also applied it on the master action space.  However, to make this efficient, the \acrshort{nn} architecture was redesigned.

\begin{figure}
\centering
	\graphicssss{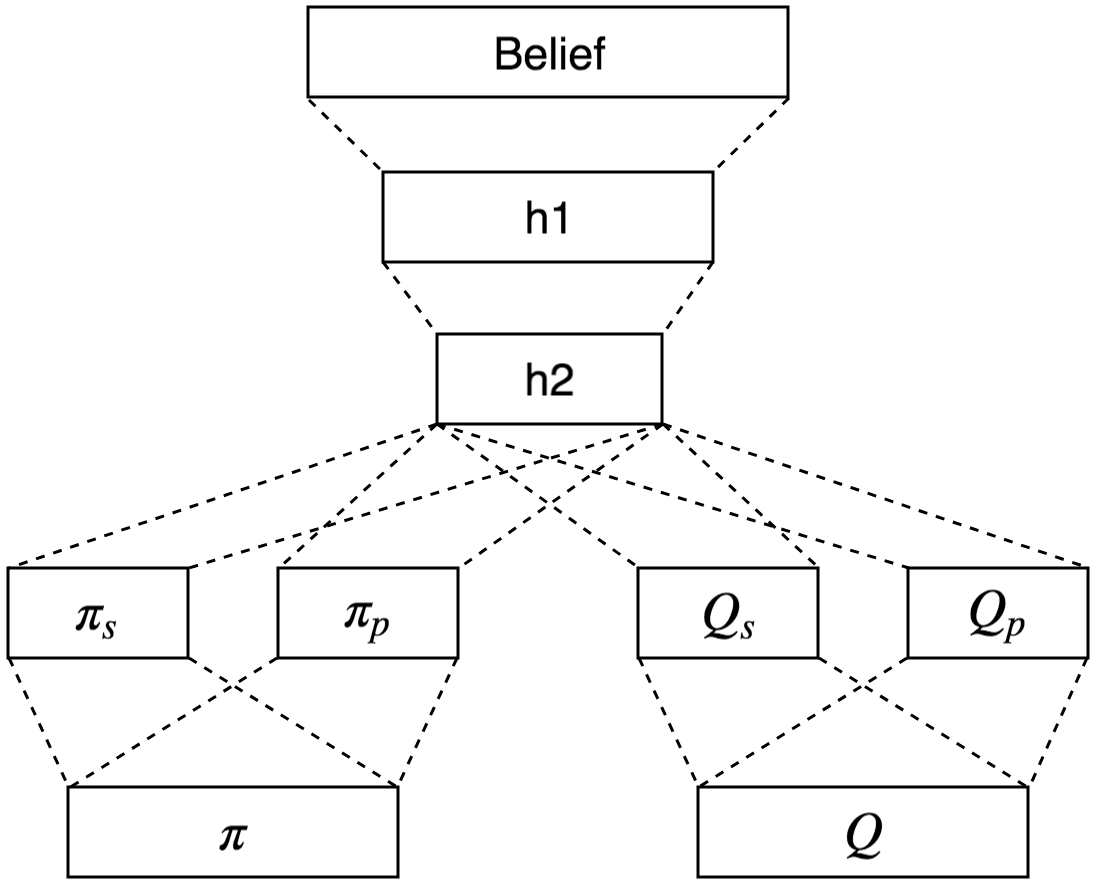}
\caption{Architecture of the actor-critic neural network for the master action space.}
% todo make smaller and maybe left-to-right
\label{macer-arch}
\end{figure}

In the case of the CamInfo domain (see Appendix~\ref{caminfo}), there are 8 informable slots of an entity, each with a binary choice on whether we inform on it. Thus, a single inform action makes up $2^8=256$ separate master actions, only differing in what they inform on. We want to incorporate the fact that these actions are very similar into the design of the \acrshort{nn} architecture. We achieve this by breaking the policy $\pi$ into a \emph{summary} policy $\pi_s$, corresponding to the 15-dimensional summary action space, and a \emph{payload} policy $\pi_p$, corresponding to the $256$ choices of the \emph{payload} of an inform action. We break the $Q$ function up similarly into $Q_s$ and $Q_p$. 
We reconstruct the 1035-dimensional master policy $\pi$ (see Appendix~\ref{caminfo}). and master $Q$-function $Q$ as follows: for each summary action $A$,
\begin{itemize}
\item If $A$ does not have a payload (i.e.\ is not an inform action), append the corresponding summary values from $\pi_s$ and $Q_s$ onto $\pi$ and $Q$.
\item Otherwise, for each payload $P$ of the $256$ possible choices, append $\pi_s(A)\cdot \pi_p(P)$ to $\pi$. This is because the probability of choosing action $A$ with payload $P$ is modelled as the product of the probability of choosing $A$ and that of choosing $P$. For each $P$, we also append $Q_s(A) + Q_p(P)$ to $Q$, allowing the payload network to learn an offset of $Q$ achieved by choosing a particular payload.
\end{itemize}
The complete \acrshort{nn} architecture is illustrated in Figure~\ref{macer-arch}. It is important to note that only the architecture of the \acrshort{nn}s is changed and the training algorithm is unchanged. In fact, the \acrshort{nn}s are treated as a black box by \acrshort{acer}. These output is a 1035-dimensional vector for master action space.

\subsubsection{Master actions for GP-SARSA}
\label{section-master-gp}

We compare \acrshort{acer} to GP-SARSA algorithm. This is an on policy algorithm that approximates the $Q$-function as a Gaussian process~(GP) and therefore is very sample-efficient~\cite{gasic2014gaussian}. The key is the use of a kernel function which defines correlations between different parts of the input space. Similarly to \acrshort{acer}, the Gaussian process~(GP) method needs to be adjusted before we deploy it on master action space. The core of a GP is the kernel function, which in the case of GP-SARSA is defined as:
$$k((b, a), (b', a')) = \langle b,b'\rangle \delta(a, a').$$
%where $b$ and $b'$ are belief states and $a$ and $a'$ are actions.
 We recall that the kernel function defines our a priori belief of the covariance between any two belief-action pairs. This kernel is a multiplication of a scalar product of the beliefs and a Kronecker delta on the actions. The latter has the effect that any two different actions are considered completely independent. While this might be a good approximation for summary actions, a more elaborate action kernel is required for master actions. This could introduce the idea that two inform actions with slightly different payloads are expected to have similar results on the same belief state, thus showing higher covariance.

Our new action kernel returns $0$ for actions $a$ and $a'$ that stem from different \emph{summary} actions. Otherwise, $a$ and $a'$ are the same inform action with differing payloads. In this case, we calculate the kernel based on the cosine similarity of the two payloads, treating the payloads as vectors describing the sets of slots to inform on. Let us call $a^s$ and $a^p$ the summary action and the payload corresponding to $a$. $a^p$ is represented as a vector where each entry is either $0$ or $1$, depending on whether the corresponding slot is informed on. Writing $\overline{a^p}=\frac{a^p}{||a^p||^\frac{1}{2}}$ for the normalised version of the payload vector, the kernel becomes $k((b, a), (b', a')) = \langle b,b'\rangle \delta(a^s, a'^s) \langle \overline{a^p}, \overline{a'^p} \rangle$. Refer to \cite{gellert} for a proof of $k$ being a valid kernel function. 
%Furthermore, the Kronecker delta function can be implemented as a scalar product between one-hot vector representations of the summary actions $a^s$. The final kernel is
%$$k((b, a), (b', a')) = \langle b,b'\rangle \langle a^s, a'^s \rangle \langle \overline{a^p}, \overline{a'^p} \rangle.$$

%To show that this is a kernel function, we show that resulting kernel matrices are positive semidefinite.
%For this, we show that for any set of $(b_i, a_i)$ pairs and vectors $x_i$ of rationals of the right dimensionality,
%$$\sum_{i,j} x_i K_{ij} x_j = \sum_{i,j} x_i k((b_i, a_i), (b_j, a_j)) x_j \geq 0$$
%This is because
%$$\begin{aligned}
%&\sum_{i,j} x_i k((b_i, a_i), (b_j, a_j)) x_j\\
%&=\sum_{i, j} \sum_k x_i x_j \cdot a^s_{ik} a^s_{jk} \cdot \overline{a^p}_{ik} \overline{a^p}_{jk}\cdot b_{ik} b_{jk}\\
% &= ||\sum_{i, k} x_i \cdot a^s_{ik}\cdot \overline{a^p}_{ik}\cdot b_{ik}||^2 \geq 0.
%\end{aligned}$$

In the case of GP-SARSA on master action space, the training algorithm is unchanged. Only the kernel function is adjusted to incorporate the idea of similarity between master actions. The \acrshort{gp} can thus be trained on the 1035-dimensional master action space.

% introduce algorithms
%  DQN not stable in dialogue
%  online vs offline algorithms
%  SARSA, IS, NAC, TRPO, A2C, PER, ..., everything
% summary vs master action space?
% pydial, user simulator, ..

%?\chapter{Related Work}

%\chapter{Design and Implementation}

\section{Limitations}
It is important to highlight some limitations of this work. This work is not addressing the problem of modelling policy with large action space where there are no similarities between the system actions. On the contrary, we focus on large action spaces where we can establish some relations between the actions, either by sharing the weights in the neural network architecture as in Section~\ref{section-macer} or by defining special kernel functions as in Section~\ref{section-master-gp}. Although, this might seem limiting, in practice, in any task-oriented dialogue, actions will bear a lot of similarities. This used to be addressed by producing a smaller summary space of distinct actions, but we believe that the proposed approach scales better, removes hand-crafting and leads to the better performance. The latter hypothesis is investigated in the next section. 

\section{Evaluation}\label{eval}

In this section, we evaluate the performance of \acrshort{acer} incorporated in an \acrshort{sds}. We find that \acrshort{acer} delivers the best performance and fastest convergence among the compared \acrshort{nn}-based algorithms (\acrshort{enac} and \acrshort{a2c}) implemented in the PyDial dialogue toolkit~\cite{ultes2017pydial}. %As \acrshort{acer} is an elaborate algorithm with many modular steps aimed at improving its convergence properties, we follow up this ovservation with an investigation of the contribution of each step.
%We follow up this observation with an investigation of the contribution of TRPO.
We also deploy the algorithm in a more challenging setting without the \emph{execution mask} aiding action selection.
Next, we investigate the effect of different hyperparameter selections, and the algorithm's stability against it. 
Then, we deploy \acrshort{acer} and GP on master action space. Finally, we investigate how resilient different algorithms are to semantic errors and changing testing conditions.

\subsection{Evaluation set-up}

%During training, the algorithm explores the state space. The exploration is generally decreasing and the algorithm exploits what it learned more and more. 
%In testing, we aim to measure the performance of the algorithm \emph{as if it stopped training} after a certain number of episodes. Thus, 
We compare our implementation of \acrshort{acer} two \acrshort{nn}-based algorithms, namely \acrshort{enac} (\cite{peters2006policy}) and \acrshort{a2c} and to a non-parametric algorithm \acrshort{gp}.

Experiments are run as follows. First, the total number of \emph{dialogues} or \emph{iterations} ($4000$) is broken down into \emph{milestones} ($20$ milestones of $200$ iterations each). As the training over the total number of iterations progresses, a snapshot of the state of the training (all \acrshort{nn} weights, hyperparameters, and replay memory) is saved at each milestone. A separate run of 4000 iterations is then performed without any training steps, where each of the saved snapshots are tested for 200 iterations. No training and no exploration is being performed during the testing phase; instead of $\epsilon$-greedy, the greedy policy with respect to $\pi$ is used to derive the next action. This informs us on the performance of the system \emph{as if it stopped training} at a specific milestone, allowing us to observe the speed of convergence and the performance of early milestones, discounting for the exploration.

We run the evaluation $15$ times and average the results, to reduce the variance arising from different random initialisations. We compare the average per-episode reward obtained by the agent, the average number of turns in a dialogue and the percentage of successful dialogues.
The reward is defined as $20$ for a successful dialogue minus the number of turns in the dialogue. The number of maximum turns is limited to $25$, after which, if the user did not achieve their goal, the dialogue is deemed unsuccessful. The discount factor $\gamma$ is set to $0.99$ for all algorithms where it is applicable. For \acrshort{nn}-based algorithms, the size of a \emph{minibatch}, on which the training step is performed, is $64$. For algorithms employing \acrlong{er}, the replay memory has a capacity of $2000 $ interactions. For \acrshort{nn}-based algorithms, $\epsilon$-greedy exploration is used, with $\epsilon$ linearly reducing from $0.95$ down to $0$ over the training process. 

 %For a successful dialogue during which the simulated user achieved their goal, there is a reward of $20$. The algorithms optimise this reward metric, rather than the success rate.

\subsection{User simulator}
We use the agenda-based user simulator, with the \emph{focus} belief tracker for all experiments. For details, see~\cite{ultes2017pydial}. The agenda-based user simulator~\cite{stwy07} consists of a goal which is a randomly generated slot-value pairs that the entity that the user seeks must be satisfied and an agenda which is a dynamic stack of dialogue acts that the user elicits in order to satisfy the goal. The simulated user consist of deerministic and stochastic decisions which govern its behaviour capable of generating complex behaviour. A typical dialogue starts by user expressing what it is looking for, or waiting for the system to prompt it. Then it checks whether the offered entity satisfies all the constraints. In that process it sometimes changes its goal and asks for something else, making it more difficult for the system to satisfies its goal. Once it settles on the offered entity, it asks for additional information, such as \emph{address} or \emph{phone-number}. For a dialogue to be deemed successful, the offered entity needs to match the last user goal. Also, the system must provide all further information that the user simulator asked for. The reward is delayed and only given at the end of the dialogue. No reward is given for partially completed tasks.

\begin{figure}[!t]
\centering
	\graphicsssss{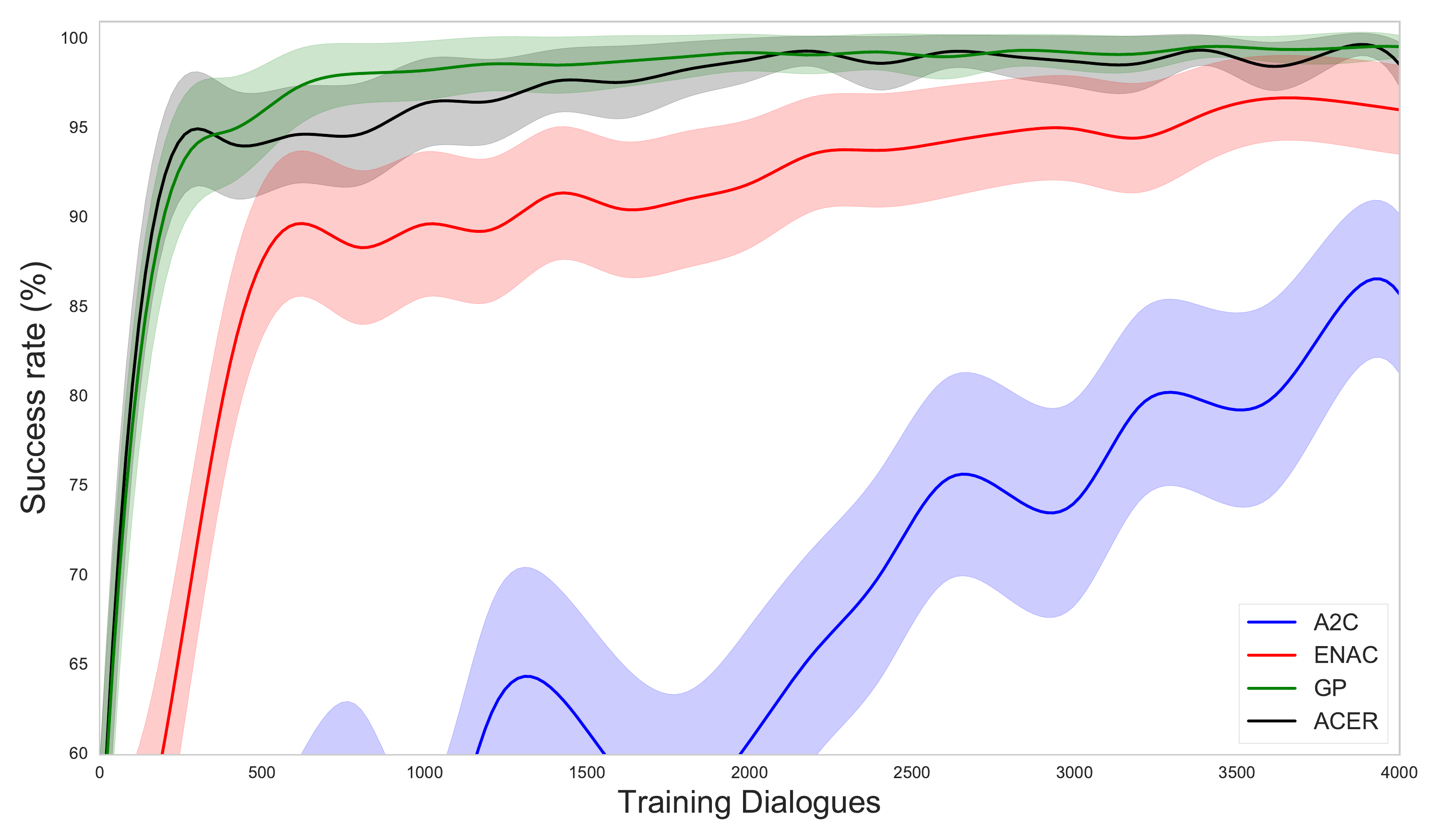}
\caption{Success rate of \acrshort{acer} compared to other \acrshort{rl} methods. Shaded areas represent a 95\% confidence interval.}
\label{plot-poster}
\end{figure}

\begin{figure}[!t]
\centering
	\graphicsssss{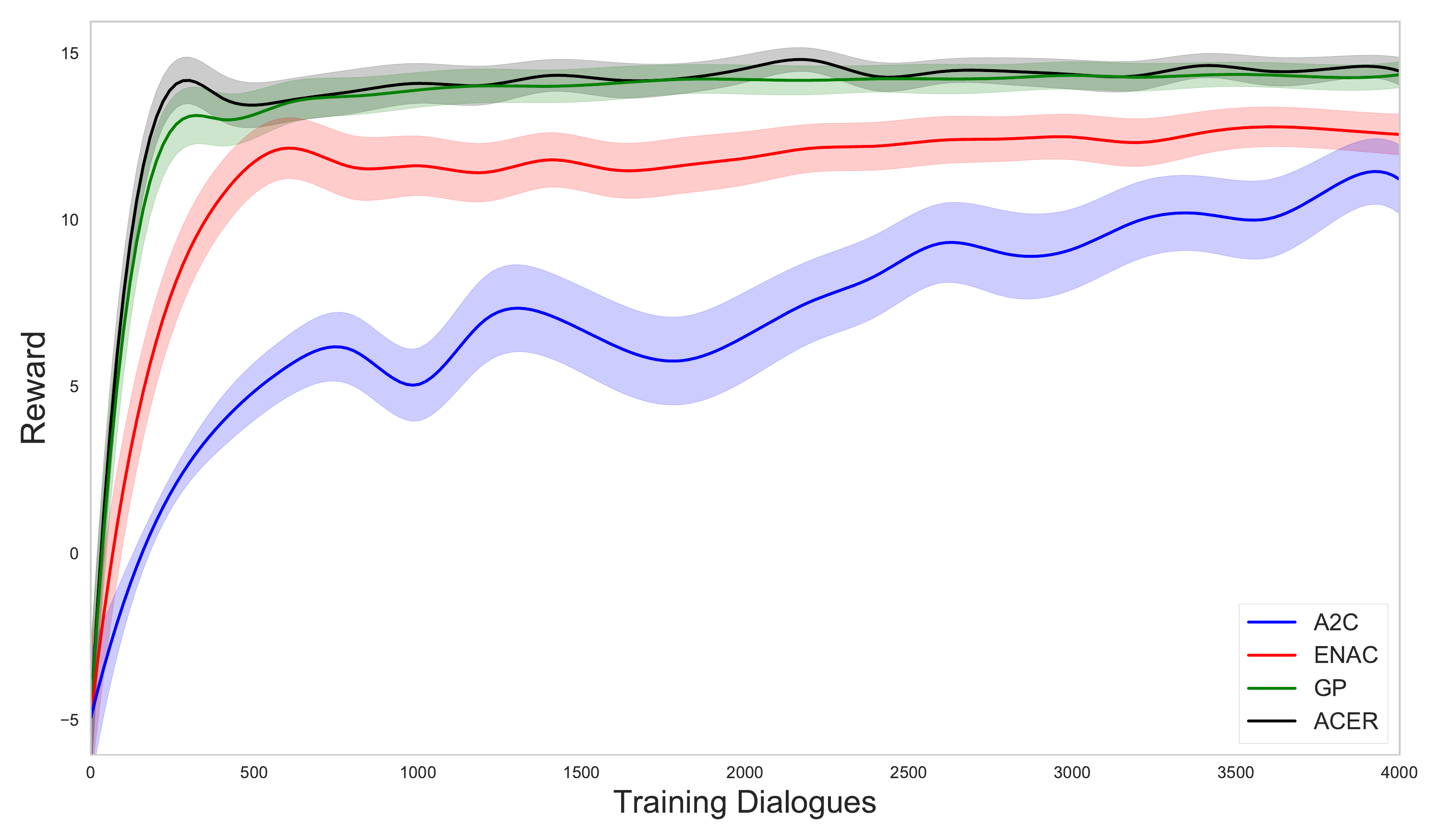}
\caption{Rewards of \acrshort{acer} compared to other \acrshort{rl} methods. Shaded areas represent a 95\% confidence interval.}
\label{plot-poster-rewards}
\end{figure}

\begin{figure}[!t]
\centering
	\graphicsssss{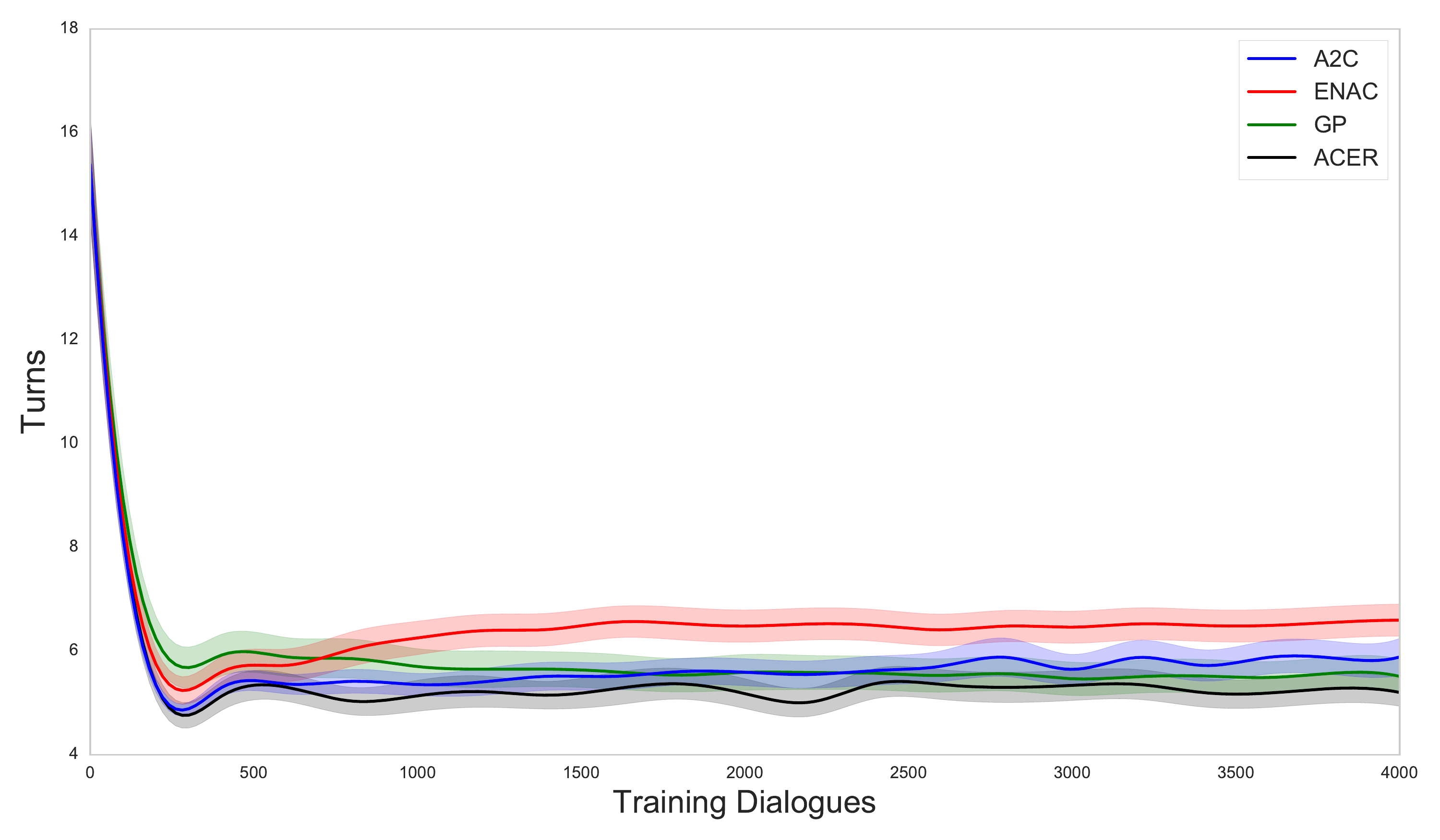}
\caption{Number of turns for \acrshort{acer} compared to other \acrshort{rl} methods. Shaded areas represent a 95\% confidence interval.}
\label{plot-poster-turns}
\end{figure}

\subsection{Performance of \acrshort{acer}}
In the initial environment, the simulated semantic error rate is 0\% both for training and testing. The learning rate $\alpha=0.001$. Instead of a simple gradient descent on the loss function, we use the Adam Optimiser, which associates \emph{momentum} to the gradient \cite{kingma2014adam}. To discourage the algorithm from learning a trivial policy, we subtract 1\% of the policy entropy from the loss function. %This way, saying eg. \emph{bye} every time is discouraged, as the entropy of such a policy would be low, thus the loss would be higher. 
The \acrshort{acer}-specific hyperparameters are: $c=5, \delta=1, \beta=0.99, n=1$. %We perform training and testing on 4000 iterations split into 20 steps. 
The results are given in Figure~\ref{plot-poster} and Figure~\ref{plot-poster-rewards} where the shaded are represents a $95\%$ confidence interval. %It is thus important to include it in the comparisons. As the training progresses, the success rate (Figure~\ref{plot-poster}) and the average reward increase (Figure~\ref{plot-poster-rewards}).

We observe that \acrshort{acer} is comparable to \acrshort{gp} in terms of speed of convergence, sample efficiency, success rate, rewards and turns. While the success rate of \acrshort{acer} remains one or two percentage points below that of \acrshort{gp}, \acrshort{acer} requires fewer dialogue turns and ultimately obtains somewhat higher rewards than \acrshort{gp}. This suggests that the slightly worse success rate of \acrshort{acer} presents a shortcoming of the reward function rather than the algorithm, as the algorithm only optimises the reward function. We also observe that \acrshort{acer} far exceeds the performance of other \acrshort{nn}-based methods in terms of all of speed of convergence, sample efficiency, success rate and rewards.

\subsection{Effect of execution mask}

\begin{figure}
\centering
	\graphicsssss{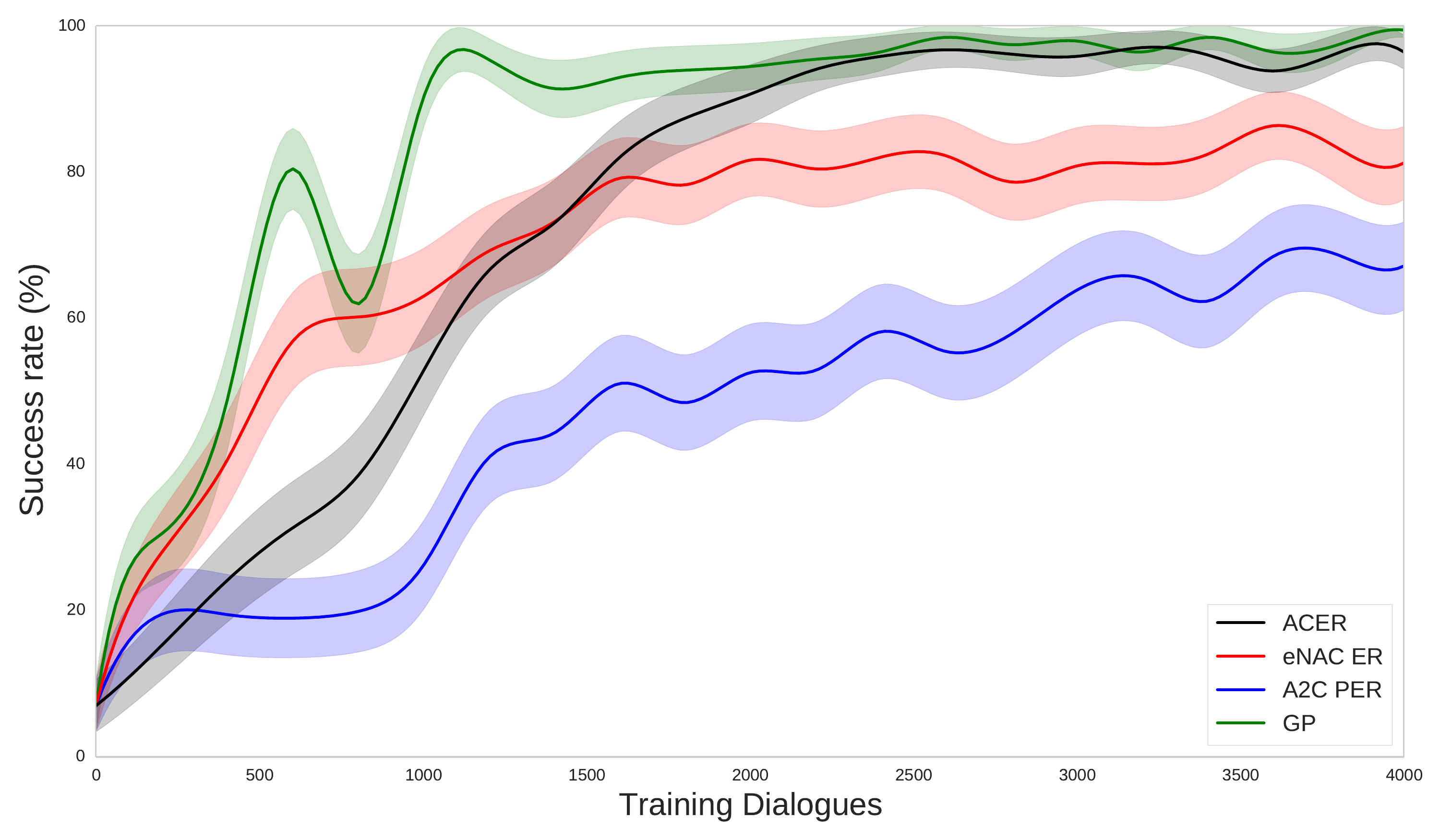}
\caption{Success rate of \acrshort{acer} compared to other \acrshort{rl} methods, without the execution mask. Shaded areas represent a 95\% confidence interval.}
\label{plot-nomask}
\end{figure}

%As introduced in Section~\ref{execmaskintro},  
We run our experiments with and without the execution mask and compare success rates (Figure~\ref{plot-poster} and Figure~\ref{plot-nomask}). 
%To test how \acrshort{acer} tackles the problem of removing the , we repeat our experiments without the execution mask. Figure~\ref{plot-nomask} compares success rates.
In general, as expected, algorithms converge slower without the execution mask, while the final performance of \acrshort{gp} and \acrshort{acer} remain somewhat below their performances with the mask. This is also expected as a mapping learned by \acrshort{rl} is rarely as precise as a hard-coded solution to a problem (execution mask). \acrshort{gp} shows faster initial convergence than \acrshort{acer}, as the latter shows a more steady progress without unexpected dips in performance. They remain comparable in every other regard.

\subsection{Hyperparameter tuning}
\label{hyperparams}

\acrshort{acer} has several additional hyperparameters compared to more traditional algorithms. We investigate the effect of hyperparameters $c, \delta, \beta,\mbox{ and }n$ on the algorithm's performance. To better illustrate the differences, we run the tests in a more challenging setting, without the execution mask. For every analysed parameter, we kept the rest of the hyperparameters set to values providing the best results from section $C$. 

\begin{figure}
\centering
	\graphicsssss{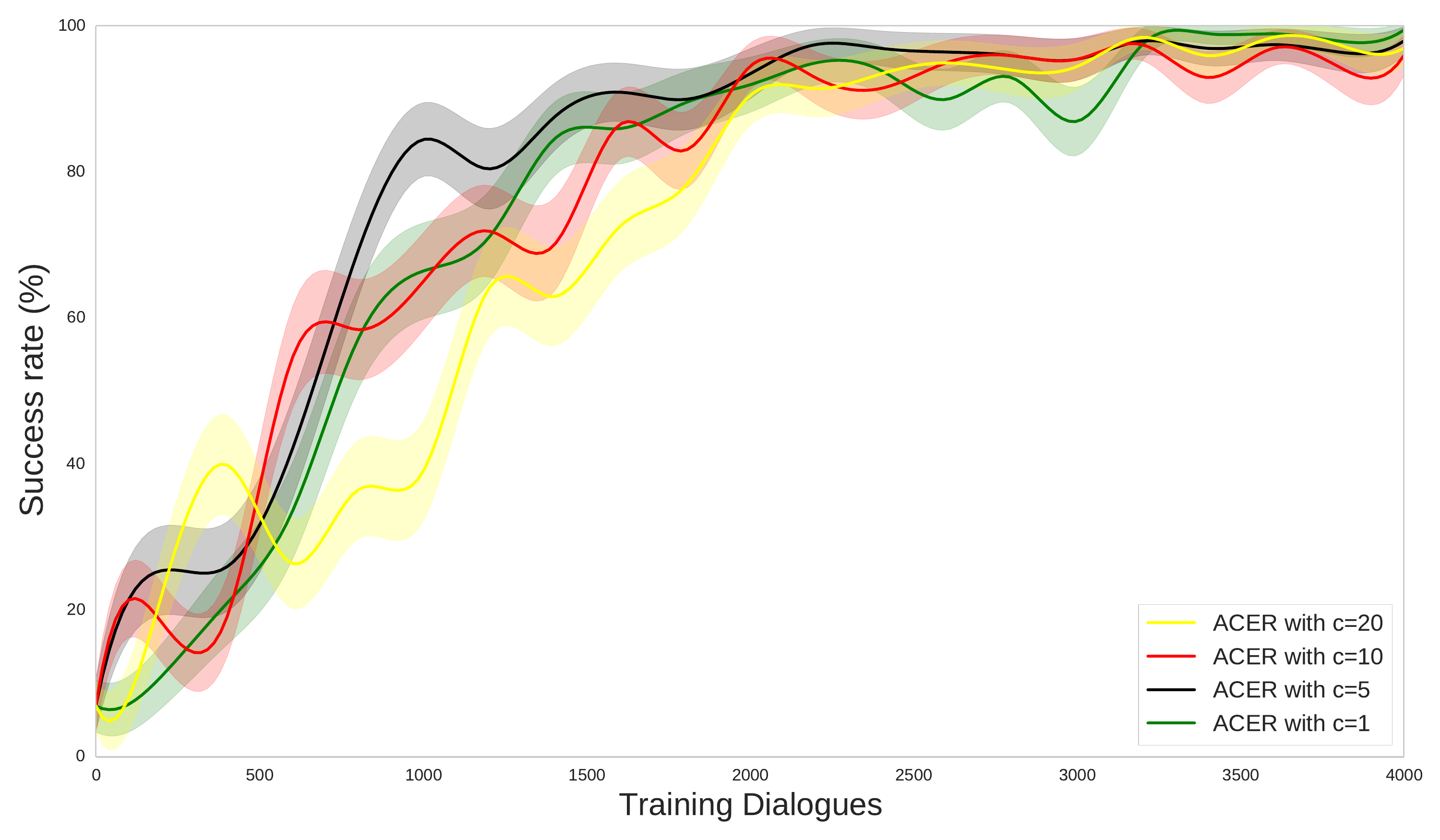}
\caption{Success rate of \acrshort{acer} with varying hyperparameter $c$. Shaded areas represent a 95\% confidence interval.}
\label{compare-c}
\end{figure}

\paragraph{\textbf{Importance Weight threshold} $c$} This value is the upper bound of \acrshort{is} weight; weights higher than $c$ are truncated. Setting this value too high diminishes the effect of weight truncation, while a value too low will rely more on the less accurate bias correction term. From Figure~\ref{compare-c}, we see that $c=5$ delivers the highest convergence rate and a good final performance. We also see that for the wide range of values from $c=1$ to $c=20$, there is no big difference in final performance, suggesting that the algorithm is relatively stable in face of varying this hyperparameter.

\begin{figure}
\centering
	\graphicsssss{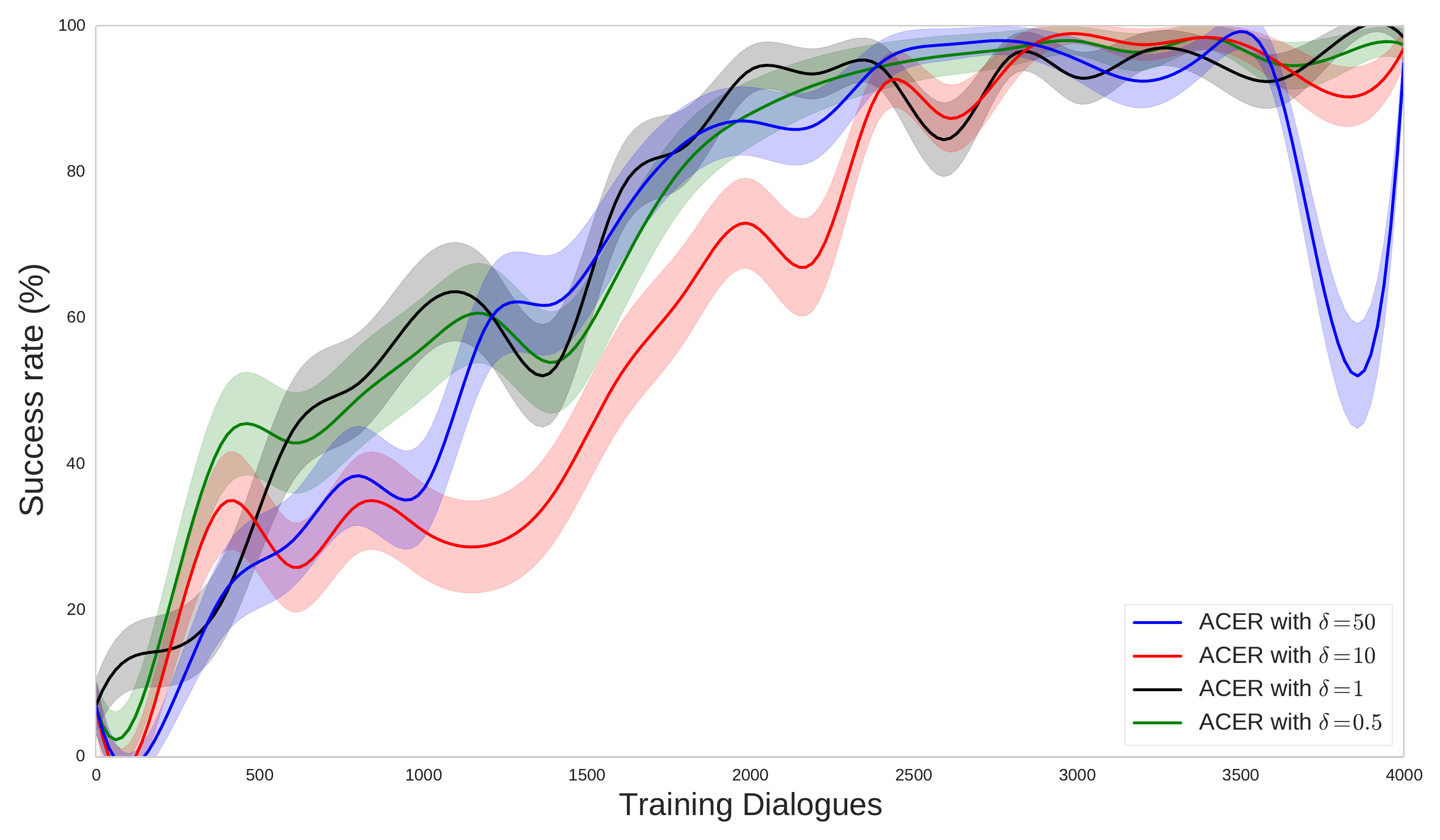}
\caption{Success rate of \acrshort{acer} with varying hyperparameter $\delta$. Shaded areas represent a 95\% confidence interval.}
\label{compare-delta}
\end{figure}

\paragraph{\textbf{KL divergence constraint } $\delta$} This value constrains the KL divergence between an updated policy and the running average policy. Setting it too high allows radical jumps, setting it too low slows the convergence down (Figure~\ref{compare-delta}). We can see that a setting of $\delta=10$ or $\delta=50$ results in erratic changes in the performance of ACER, while $\delta=0.5$ and $\delta=1$ are sensible choices.

\begin{figure}
\centering
	\graphicsssss{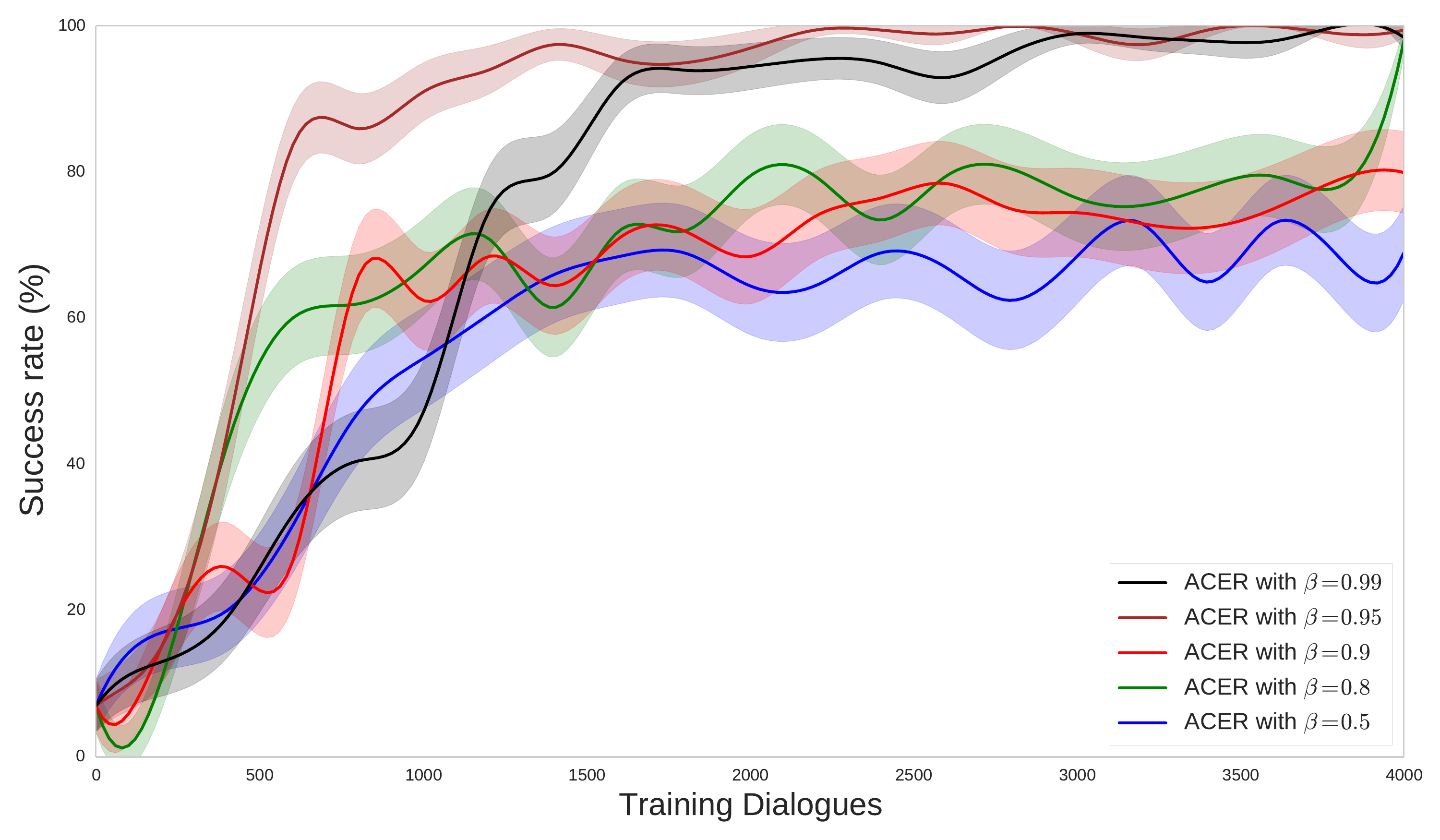}
\caption{Success rate of \acrshort{acer} with varying hyperparameter $\beta$. Shaded areas represent a 95\% confidence interval.}
\label{compare-beta}
\end{figure}

\paragraph{\textbf{Average policy update weight} $\beta$} %As discussed in Section~\ref{section-beta}, a too low value makes the average policy forget the history very quickly, but a too high value will prevent the average policy to adjust to the current policy. Figure~\ref{compare-beta} shows the comparisons. 
In Figure~\ref{compare-beta}, we can see that for $\beta\leq0.9$, the average policy forgets the history too quickly, allowing the policy to gain momentum in any direction and thus preventing it from converging to a good performance. For $\beta=0.95$, the policy converges quickly, while $\beta=0.99$ results in a somewhat conservative algorithm, where the KL divergence constraint keeps the policy near a slowly changing average. $\beta=0.99$ still converges to a good result, but does so somewhat slower than in case of $\beta=0.95$.

\begin{figure}
\centering
	\graphicsssss{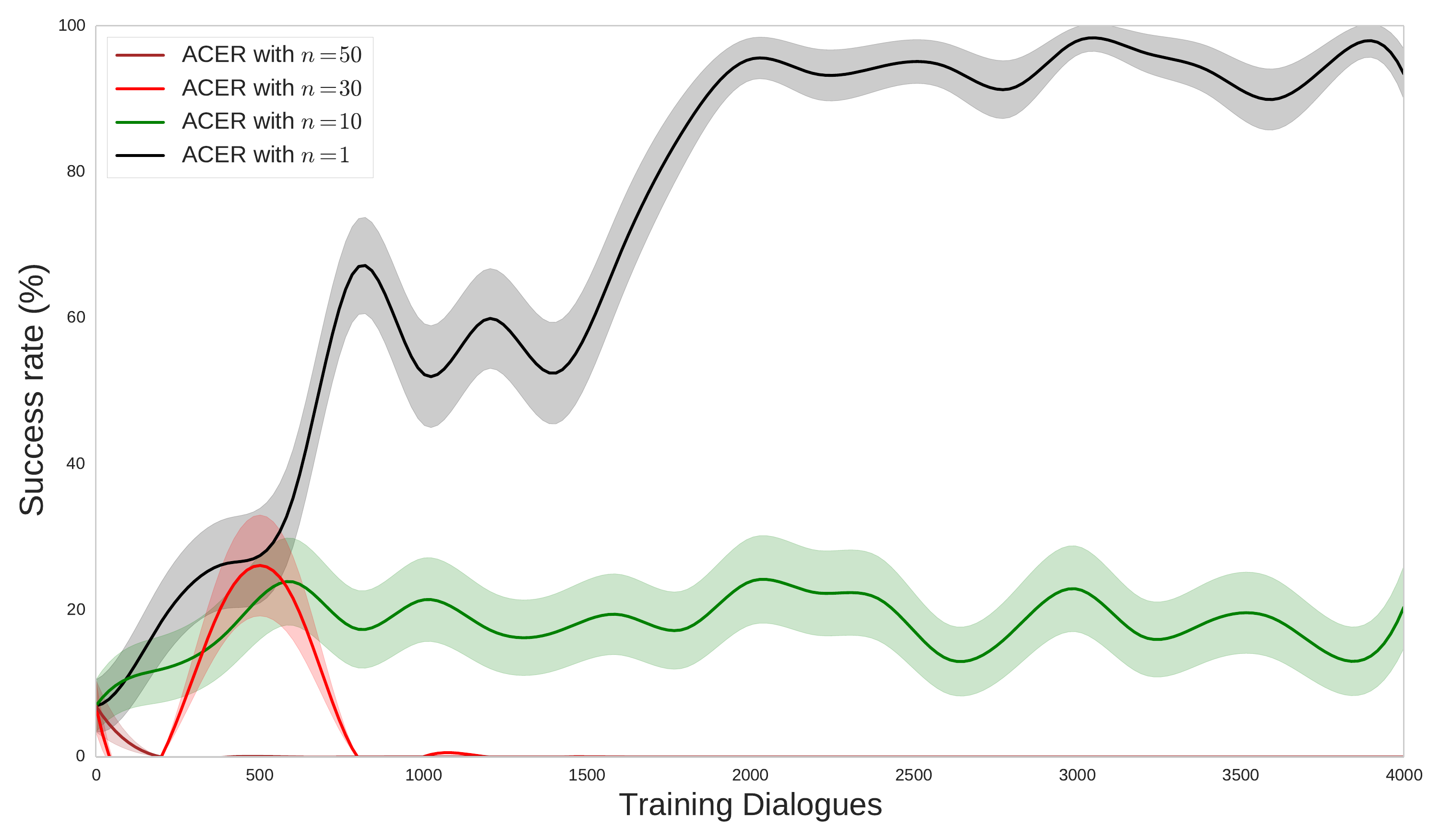}
\caption{Success rate of \acrshort{acer} with varying hyperparameter $n$. Shaded areas represent a 95\% confidence interval.}
\label{compare-iters}
\end{figure}

\paragraph{\textbf{Training iterations} $n$} %For each episode gathered, we run the training step $n$ times. 
Setting the number of training steps per episode $n$ higher allows the algorithm to learn more from the gathered experience. However, if $n$ is too high, the training might diverge due to the policy moving too much (Figure~\ref{compare-iters}). For $n=1$, convergence is quick and performance is good, while for $n=10$, performance stays poor throughout. For $n=30$ and $n=50$, the algorithm diverges completely.

\subsection{Master action space}

ACER compares favourably to other \acrshort{nn}-based algorithms, but performs about equally if not slightly worse than GP in our experiments. 
The experiments were run on the summary action space, which only has $15$ actions. In a more difficult scenario, we may have orders of magnitude more actions. In such scenarios, the computational cost of GPs can be prohibitive as it needs to invert the Gram matrix~\cite{gasic2014gaussian}. If ACER still performs well under the same scenario, it might may be the overall best method to apply to larger action spaces. This is because ACER does not have the prohibitive  computational cost of GP, and is expected to train much more quickly.

\begin{figure}
\centering
	\graphicsssss{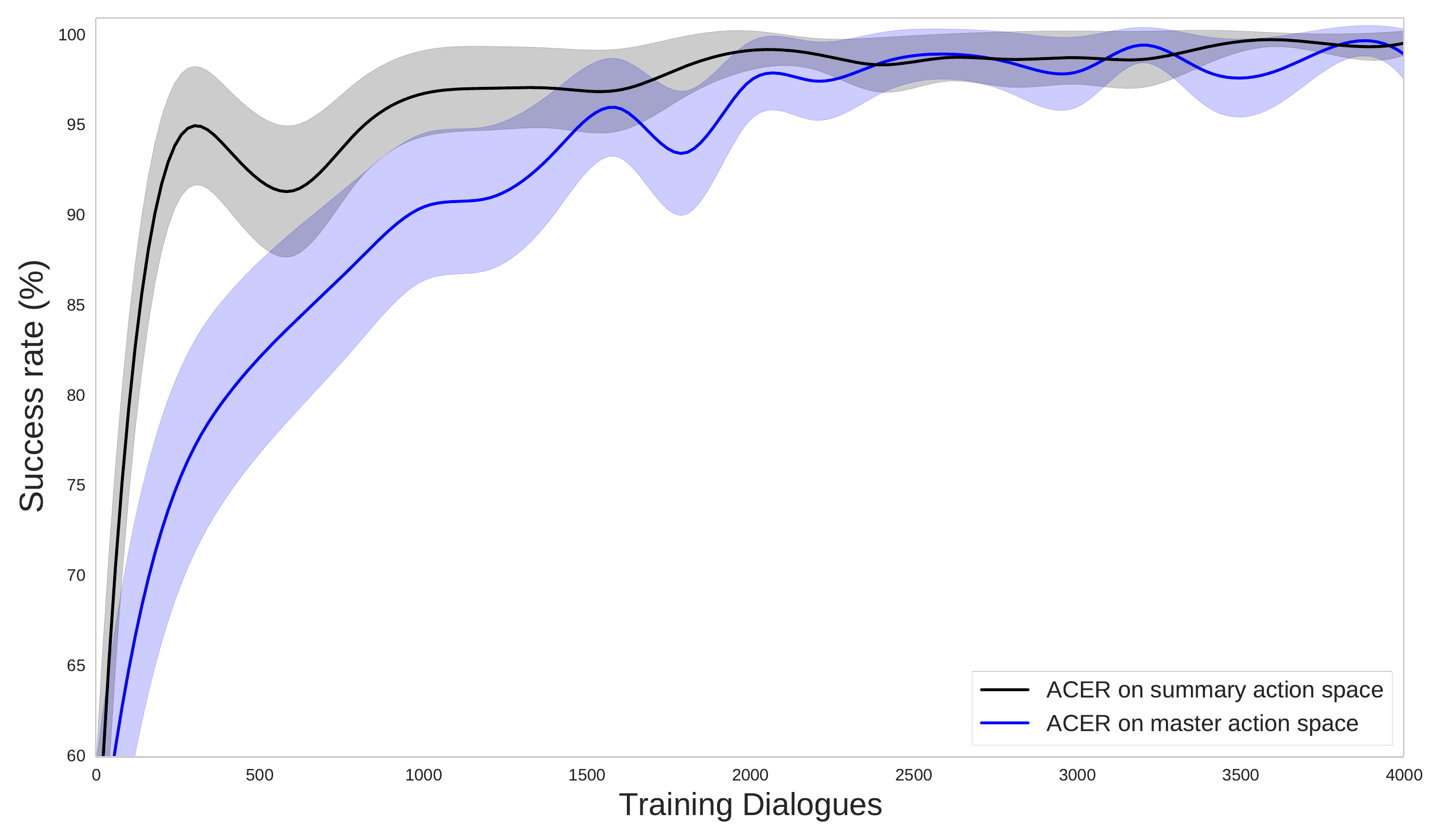}
\caption{Success of ACER on summary action space compared to ACER on master action space with execution mask. Shaded areas represent a 95\% confidence interval.}
\label{compare-master}
\end{figure}

To test our hypotheses, we deploy ACER on the master action space according to Section~\ref{section-macer}, and on the summary space (Figure~\ref{compare-master}). Both experiments were run with the execution mask. Convergence is slower on the master action space. This is expected due to having to choose between vastly higher number of actions on the master action space ($1035$ as opposed to $15$). However, ACER is still surprisingly effective on the master action space, converging to about the same performance as on the summary space. We note that this is without any modification to the training algorithm, %; as described in Section~\ref{section-macer}, 
only the underlying \acrshort{nn} is changed. ACER achieves the best results in terms of speed of convergence and final performance on master action space out of NN-based SDS policy optimiser algorithms.

\begin{figure}
\centering
	\graphicsssss{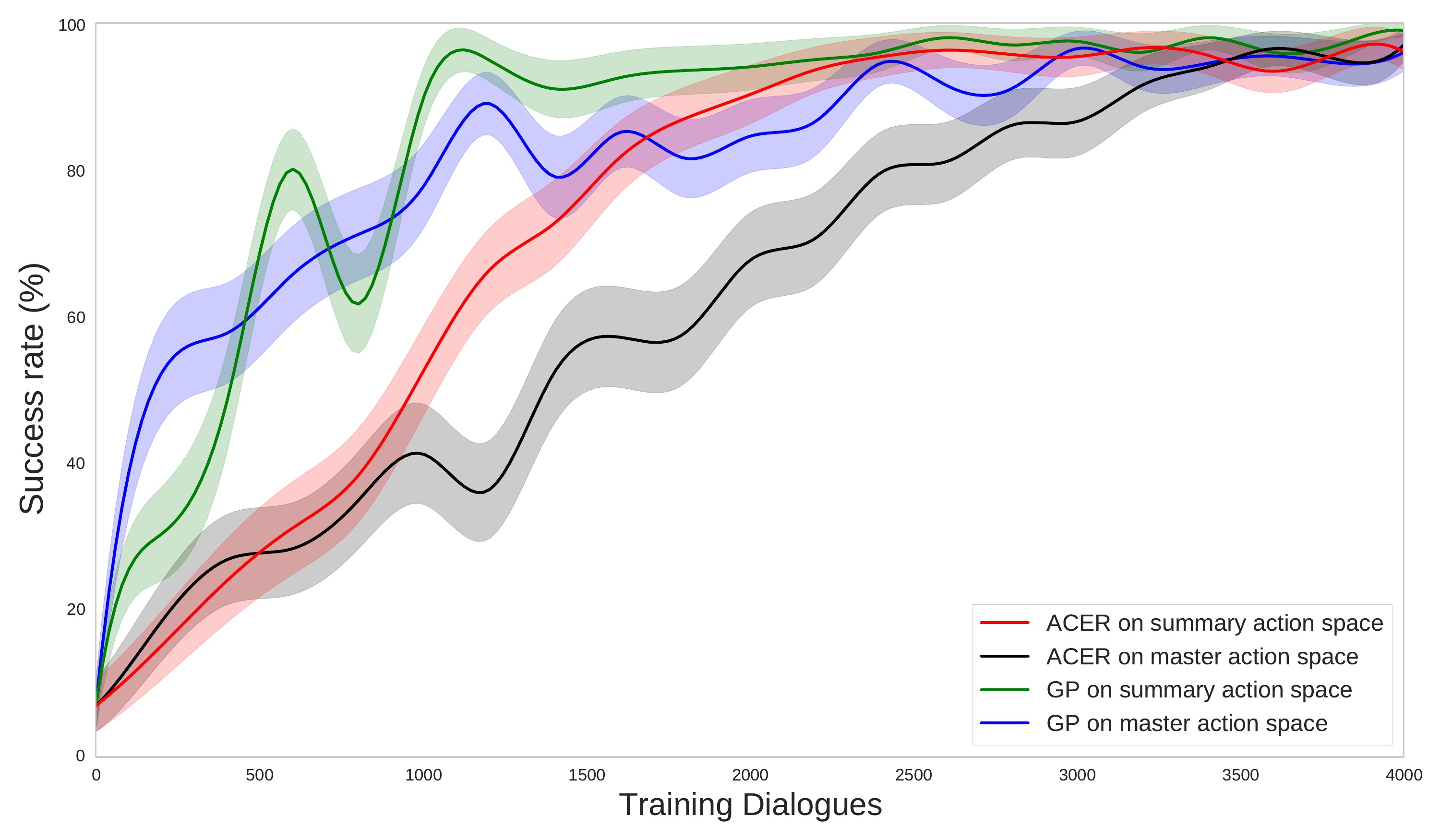}
\caption{Success of ACER and GP on summary and master action spaces, without execution mask. Shaded areas represent a 95\% confidence interval.}
\label{plot_master_gp_nomask_success}
\end{figure}

To investigate further whether ACER is the best choice of algorithm on the master action space, we modify GP to run on master action space according to Section~\ref{section-master-gp}. We compare ACER and GP both on summary and master action spaces, without the execution mask in Figure~\ref{plot_master_gp_nomask_success}. Both GP and ACER show slower speed of convergence on master action space. This is expected, as the random initialisation of a policy on master action space will be much less sensible than an initialisation on the summary space, the latter taking advantage of the hard-coded summary to master action mapping method.
However, it is surprising to see that all experiments converged to roughly the same performance of about 97\% success rate, except for GP on summary, which has a final success rate of 98\%-99\%. This suggests that both ACER and GP can handle large action spaces quite efficiently. To our knowledge, this is the first time learning on the master action space from scratch was successfully attempted.
%To our knowledge, our implementation of both ACER and GP on master action space achieves better performance than any experiment of a policy optimiser on master action space in SDS.

GP is more sample efficient than ACER on the challenging master action space without execution mask. However, it requires vastly more computational resources to run: this experiment took $6.45$ hours to run with ACER, and $8.63$ days with GP\footnote{The running times were measured on an Azure cloud machine with a 16-core CPU and 64GB of RAM.}. Arguably, the extra computational cost overshadows the disadvantage of ACER, that it has to be run for more iterations to converge. 

% nomask master
% gp: 12429 minutes
% macer: 387 minutes

%\section{\acrshort{gp} on master action space}

\subsection{Noise robustness}
\label{errors}

So far, our experiment settings were quite idealised, training and testing policies under a perfect simulator with no semantic errors. However, in real life the automatic speech recognition~(ASR) component is very likely to make errors as well as the spoken language understanding~(SLU) component. Therefore, in reality, the pipeline surrounding the policy optimiser deals with substantial uncertainty, which tends to introduce errors~\cite{lcgc17}. We ultimately want to measure how well a policy optimiser can learn the optimal strategy in face of noisy \emph{semantic}-level input. In our experiments, we control this by the \emph{semantic error rate}, the rate at which a random noisy input is introduced to the optimiser to simulate an error scenario. In other words, a 15\% semantic error rate means that with 0.15 probability a semantic concept (slot, value or dialogue act type) presented to the dialogue manager is incorrect.  We focus on two desirable properties of a policy. First, ideally, the policy would learn not to \emph{trust} the input as much, and ask questions until it is sure about the user goal, just like a real human would if the telephone line is noisy. Second, an ideal policy would not only adjust to the error rate of the training conditions, but would dynamically adjust to the conditions of the dialogue it is in. If the policy adjusts too much to the training conditions, it is said to \emph{overfit}. This could severely limit the policy's deployability.

\begin{figure}
\centering
	\graphicsssss{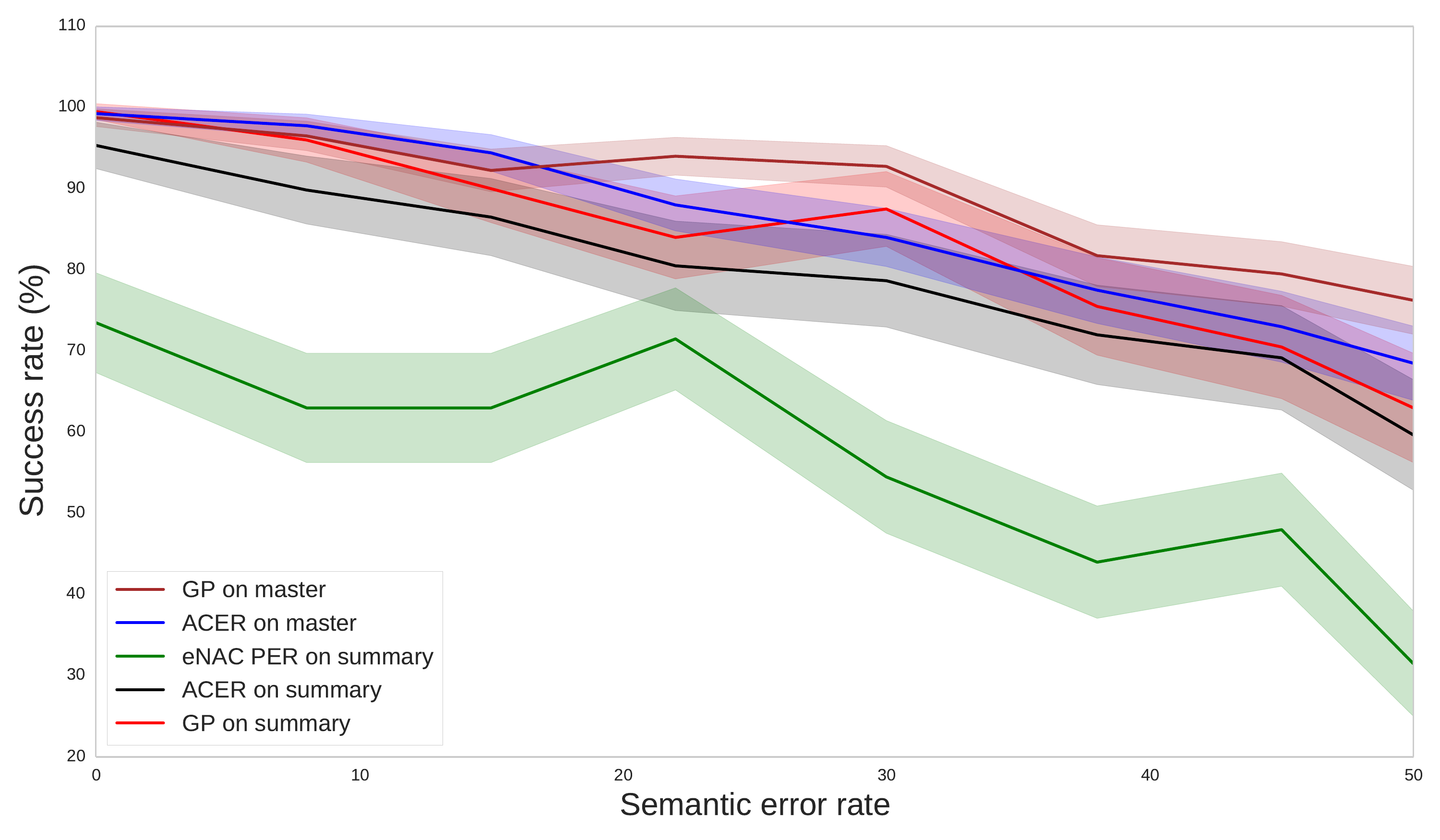}
\caption{Success rate of key algorithms when training them on 15\% and testing them on varying error rates. Shaded areas represent a 95\% confidence interval.}
\label{plot2}
\end{figure}

\begin{figure}
\centering
	\graphicsssss{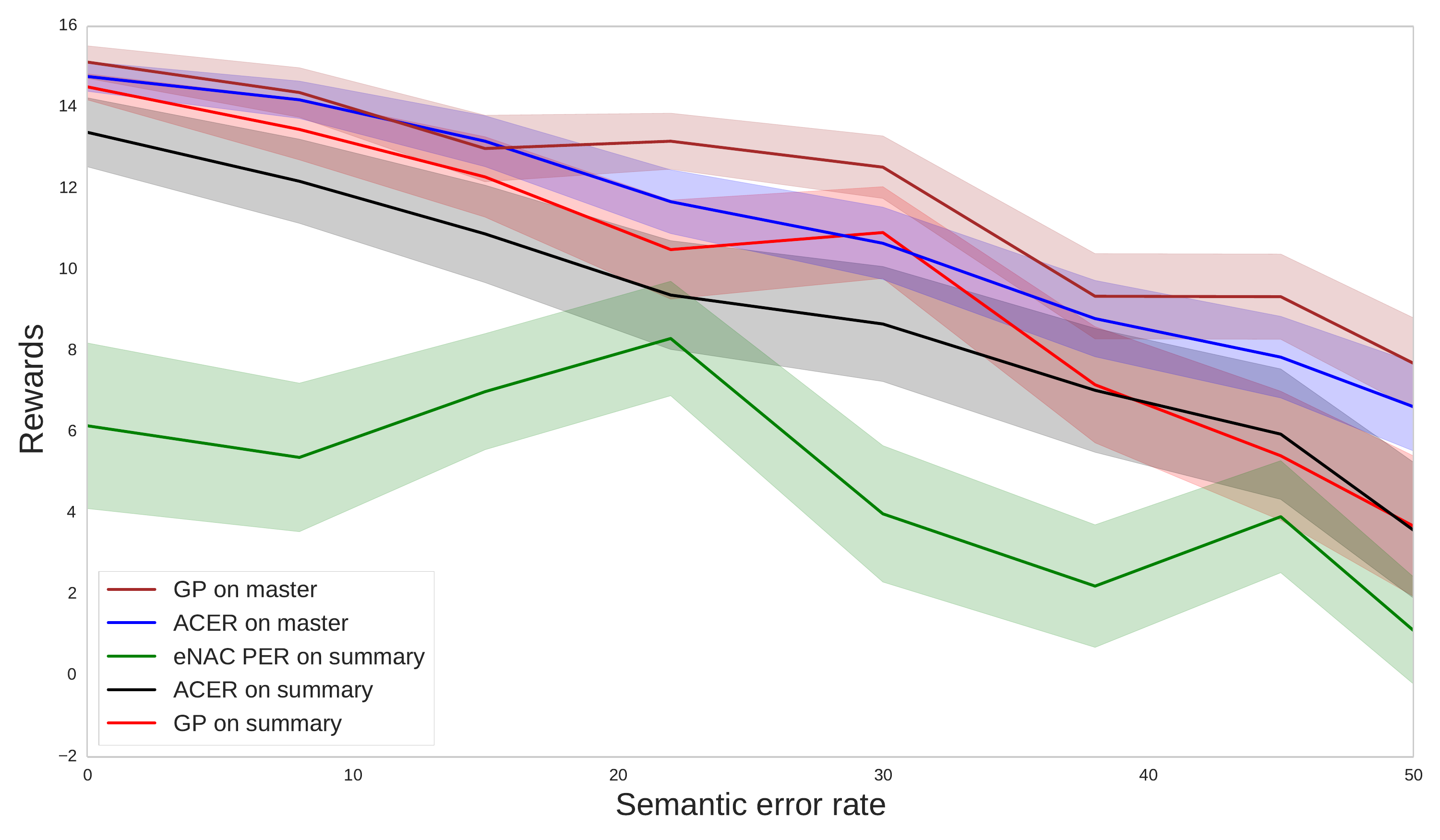}
\caption{Rewards of key algorithms when training them on 15\% and testing them on varying error rates. Shaded areas represent a 95\% confidence interval.}
\label{plot2_rewards}
\end{figure}

We test key algorithms for these two desirable properties. eNAC, the best known NN-based policy optimiser~\cite{su2017sample} to this date, is compared to ACER and GP. ACER and GP are also compared to their respective variants in master action space. We run the test as follows: first, we train the algorithms under $15\%$ semantic error rate until convergence, with the execution mask. Then we take the fully trained policy and test it under a range of semantic error rates, ranging from $0\%$ to $50\%$ to measure the policies' generalisation properties. This is something that is never the case in games so this aspect of learning is rarely examined but it is of utmost importance for spoken dialogue systems. We present results in Figure~\ref{plot2} and Figure~\ref{plot2_rewards} with a 95\% confidence interval.


Success rate and reward follow the same trends. 
As expected, we see a general downwards trend for each algorithm as the semantic error rate increases. 
%Due to the computational cost of this experiment, some algorithms were only run once
There is however no apparent spike in performance at the 15\% semantic error rate of the training process, indicating that none of the algorithms overfit to this setting. We can see that the performance of eNAC is far behind all the other algorithms. ACER and GP are closer in performance, but GP on summary space consistently beats ACER on summary space. 

It might be surprising that both ACER and GP perform better when trained on the master action space as opposed to the summary space, given that they performed worse in previous experiments. However, those experiments had no semantic errors, and a hand-crafted rigid mapping from summary actions to master actions, that relied on the belief state to find the best payload for an inform action. Under a higher semantic error rate, the belief state will be noisy and this mapping may not perform optimally. This highlights the benefits of expanding the scope of artificial intelligence in SDS: AI can be more versatile than hand-coded mappings, especially when the mapping performs \emph{decision making under uncertainty}.

\subsection{Human evaluation}
\label{mturks}
In the previous sections, the training and testing is performed on the same simulated user. To test the generalisation capabilities of the proposed methods, we evaluate the trained dialogue policies in interaction with human users in a similar set-up as in~\cite{jkgm11}. To recruit the users, we use the Amazon Mechanical Turk (AMT) service where volunteers can call our dialogue system and rate it. Around $900$ dialogues were gathered. 
Three policies (GP and ACER on summary action space and ACER on master action space) were trained with $15\%$ semantic error rate to accommodate for ASR errors using set-up from previous sections. Then, learnt policies were incorporated into SDS pipeline with a commercial ASR system.

The MTurk users were asked to find restaurants that have particular features as defined by the given task. Subjects were randomly allocated to one of the three analysed systems. After each dialogue the users were asked whether they judged the dialogue to be successful or not which is then translated to a reward measure. Table \ref{mturks} presents averaged results with one standard deviation. All models differ indiscernibly with regards to success rate performing very well. However, ACER trained on master action space achieves considerably higher reward (and in turn smaller number of turns) than models working on summary action space.  
\begin{table}[!h]
\caption{Human evaluation}
\label{mturks}
\centering
\begin{tabular}{|l||c||c||c|}
\hline
 & GP summary & ACER summary& ACER master\\
\hline
Success rate & 89.7\% & 88.7\% & 89.1\% \\
\hline
Reward & 11.29 ($\pm$ 7.54) & 11.39 ($\pm$ 7.17)  & 11.83 ($\pm$ 8.05)\\
\hline
No. of turns & 6.61 ($\pm$ 3.12) &  6.42 ($\pm$ 2.84) & 5.98 ($\pm$ 3.22)\\
\hline
\end{tabular}
\end{table}

%%%%%%% CONCLUSIONS
\section{Conclusion}\label{concl}
The policy optimisation algorithms presented in this paper improves the state-of-the-art in spoken dialogue systems~(SDS) in three ways:
\begin{itemize}
\item A version of ACER \cite{main} designed for \acrshort{sds}s shows better results than the current state-of-the-art for \acrlong{nn}-based policy optimisers \cite{su2017sample}.
\item This implementation of ACER is also able to train efficiently in the master action space, showing the best performance among \acrlong{nn}-based policy optimisers, as reported by \cite{fatemi2016policy} and \cite{su2017sample}.
\item Our implementation of GP with a redesigned kernel function achieves the best performance on master action space, which previously was not possible.
\end{itemize}
GP suffers from an inherently high computational cost, making the algorithm unsuitable in higher volume action spaces. In such cases, the fact that ACER can be trained well on the master action space indicates that it may be the best currently known method to train policies with large action spaces.

As agents powered by machine learning gain more intelligence, they can be applied to more challenging domains. Using the master action space is a good example of this: a hard-coded mapping between summary and action spaces \emph{can} be used to simplify the task of the AI agent. However, as we have shown, it is no longer \emph{required} to train in this action space. There is an algorithm (ACER) that can finally bridge the semantic gap between summary and master action spaces without the help of domain-specific code written explicitly for this mapping\footnote{The design of neural networks in ACER was optimised for the dialogue management task, as described in Section~\ref{section-macer}. However, the training algorithm itself remains general.}. This has three benefits: first, training on master action space outperforms the mapping based on fixed code, when uncertainty (semantic errors) is involved. Second, it allows us to build a more generally applicable system, with less work required to deploy it in differing domains. Third, it allows us to consider domains that have vastly higher action spaces, even if there is no clear way to convert those action spaces into small summary action spaces (such as a general purpose dialogue system).

ACER fits well into other \acrshort{sds} research directions too. Successful policy optimisers need to be sample efficient and be able to be trained quickly, to avoid subjecting human users to poor dialogue performance for long. ACER uses experience replay for sample efficiency, together with many methods aimed at reducing bias and variance of the estimator, to achieve quick training.

We introduce some of the many directions in which this work could be continued.  Recently, \cite{su2017sample} combined supervised learning~(SL) with deep reinforcement learning~(DRL) to investigate the performance of an agent bootstrapped with SL and trained further with DRL. The \acrshort{nn}s of ACER are compatible with that approach. This may decrease the overall interactions required for convergence, as well as increase sample efficiency.

Both of our settings, training on summary and on master action space, considered static action spaces only. Under this framework, the entire policy would have to be retrained if a new action or payload were to be introduced. This could hurt the maintainability of a real-life dialogue system, as it would be expensive to extend the database schema or the list of actions. Ideally, the training algorithm could adapt to such changes made, being able to retain its pre-existing knowledge of the old actions and this is an important topic to investigate in the future.

% or


\appendices

\section{CamInfo action space}\label{caminfo}
We define the \emph{action space} in the CamInfo restaurants domain. Most information-seeking domains have a similar overall architecture. 
\begin{itemize}

\item \textbf{request + slot} where \emph{slot} is an informable slot such as area, food, or pricerange. This action prompts the user to specify their criteria on a slot, eg.\ \emph{``Which area are you interested in?''}

\item \textbf{confirm + slot} where \emph{slot} is an informable slot. This action prompts the user to confirm their criteria on a slot that they may or may not have already mentioned. Due to errors accumulating during the decoding pipeline (speech recognition, semantic decoding, belief tracking), the system has to deal with considerable uncertainty, but it can attempt to increase its certainty in the user's criteria by using a confirm action, eg.\ \emph{``Did you say you want an expensive restaurant?''}

\item \textbf{select + slot} where \emph{slot} is an informable slot. This action prompts a user to select a value for the slot from a specified list of values. This is less open-ended than a request action and more open-ended than a confirm action, eg.\ \emph{``Would you like Indian or Korean food?''}.

\item \textbf{inform + method + slots} action provides information on a restaurant. The associated \emph{method} specifies how the restaurant to give information on should be chosen. 
%The exact choice of restaurant is not part of the action specification; it is derived by code when converting from the action into the \emph{dialogue act}. This conversion process reads the action and the \emph{belief state} and communicates with the database in the \emph{ontology}, selecting a specific restaurant by applying heuristics.
The \emph{standard method} is to choose the first result in the \emph{ontology} that matches the user criteria specified so far. The \emph{method} can also be \emph{byname}, in which case the system believes that the user asked about a specific restaurant by referring to its name, and information on that restaurant should be provided. If the method is \emph{requested}, we inform on the same restaurant we informed on last, if it is \emph{alternatives} then we pick another restaurant that matches the user's criteria (if possible).

There are several properties of a restaurant, with a binary choice for each of them on whether the system wants to inform on it in a dialogue turn or not. The \emph{informable} slots for restaurants are: \emph{area, food type, description, phone number, price-range, address, postcode} and \emph{signature}.

We note that some of these slots are also \emph{requestable}, allowing a user to query a restaurant based on those slots. These slots are \emph{area, food type} and \emph{price-range}. % question name requestable? everything informable? name requestable tood
A restaurant also has a \emph{name}, which we will always inform on. Thus, the system has a choice between $2^8=256$ different ways it can inform on a restaurant. A specific choice is referred to as the \emph{payload} of an inform action.

\item \textbf{reqmore} is a simple action that prompts the user to provide more input.

\item \textbf{bye} is used to end the call, normally only as a response to the user's intention to end the call.

\end{itemize}

For the CamInfo domain, there are $4\cdot 2^8 = 1024$ inform actions and $3\cdot 3+2=11$ other actions, making up $1035$ actions in total. We call this action space the \emph{master action space}.  In the summary space, the inform actions do not specify which slots to inform on, leaving only $4$ separate inform actions, and $15$ actions in total.
% An example of a master action and corresponding summary action is:
% $$\begin{aligned} \mbox{Master action: }&\quad\mbox{inform}( \mbox{price}=\mbox{cheap}, \mbox{area}=\mbox{centre})\\
% \mbox{Summary action: }&\quad\mbox{inform\_requested}\\
% \end{aligned}$$
% maybe include a diagram

\section{Example dialogue}
\label{example-dialogue}

Below is an example dialogue between a user looking for a restaurant with a medium price range, and a system that internally translates between summary and master actions. System responses are written as $\mbox{Sys: } \mbox{summary\_action} \rightarrow \mbox{master\_action}$:

\begin{ssmall}
$$\begin{aligned}
Sys:&\; \mbox{hello()} \rightarrow \mbox{hello()}\\
User:&\; \mbox{inform(type=restaurant)}\\
Sys:&\; \mbox{request\_pricerange()} \rightarrow \mbox{request\_pricerange()}\\
User:&\; \mbox{inform(pricerange=moderate)}\\
Sys:&\; \mbox{inform()} \rightarrow \mbox{inform(food="turkish", }%pricerange="moderate",} 
%\\& \qquad\qquad\qquad\;
\mbox{name="efes restaurant")}\\
User:&\; \mbox{reqalts()}\\
Sys:&\; \mbox{inform\_alternatives()} \rightarrow \mbox{inform(food="turkish", }
%\\ & \qquad\qquad\qquad\;
\mbox{%pricerange="moderate", 
name="anatolia")}\\
User:&\; \mbox{request(phone)}\\
Sys:&\; \mbox{inform\_requested} \rightarrow \mbox{inform(name="anatolia",}
%\\&\qquad\qquad\qquad\;
\mbox{phone="01223 362372")}\\
User:&\; \mbox{thankyou()}\\
Sys:&\; \mbox{bye()}\\
\end{aligned}$$
\end{ssmall}

\section*{Acknowledgment}

The authors would like to thank all members of the Dialogue Systems Group for useful comments and suggestions.

\ifCLASSOPTIONcaptionsoff
  \newpage
\fi



%\bibliographystyle{IEEEtran}
\bibliographystyle{IEEEtran} % apalike
%\cleardoublepage
\bibliography{references}


%\vfill

\vspace{-2em}
\begin{IEEEbiography}[{\includegraphics[width=1in,height=1.25in,clip,keepaspectratio]{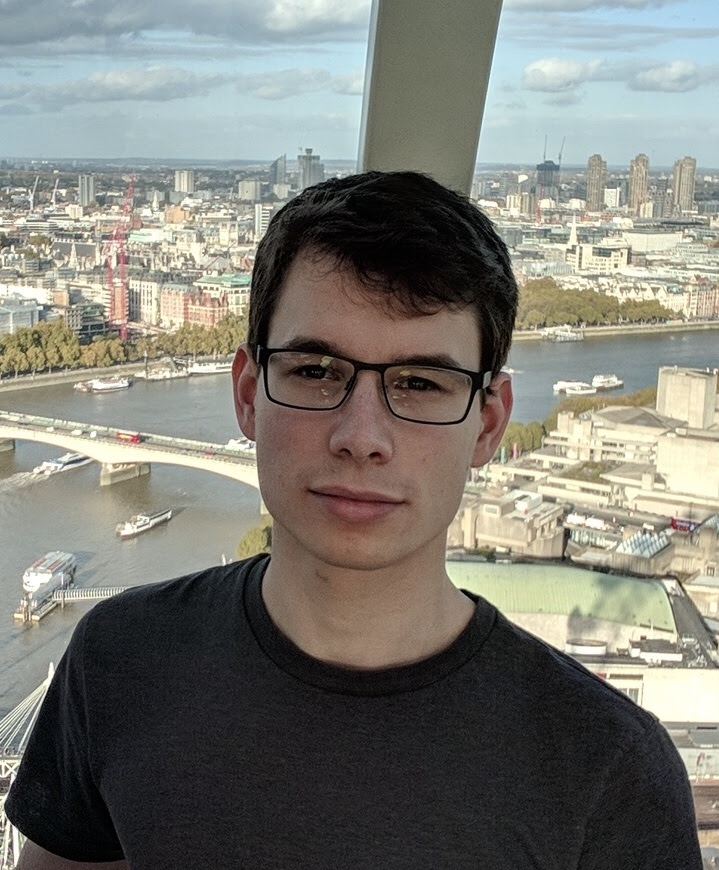}}]{Gell\'{e}rt Weisz}
%Gell\'{e}rt Weisz 
received his B.A. degree from the Computer Science Tripos at the University of Cambridge in 2016. He stayed at Cambridge as an MPhil student reading Machine Learning, Speech and Language Technology. Afterwards, he joined Deepmind as a Research Engineer.
\end{IEEEbiography}
\vspace{-2em}
\begin{IEEEbiography}[{\includegraphics[width=1in,height=1.25in,clip,keepaspectratio]{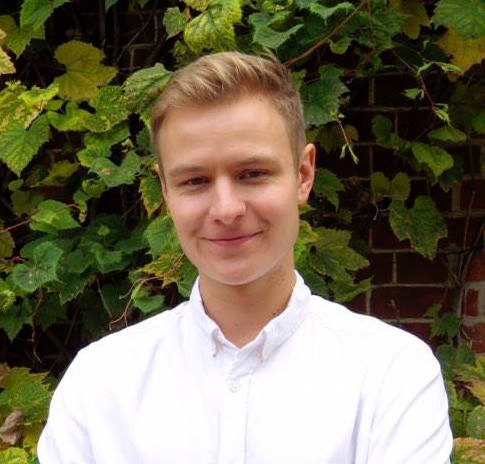}}]{Pawe\l~ Budzianowski}
%Pawe\l~ Budzianowski 
received his B.A. and M.A. degrees from the Faculty of Mathematics and Computer Science at Adam Mickiewicz University in Pozna\'n in 2015.
Since  then  he  has  been  at  the  University of  Cambridge  first  as  an  MPhil  student  reading Machine Learning, Speech and Language Technology. Afterwards, he has begun a PhD in the Dialogue Systems Group at the University of Cambridge. His research interests include multi-domain policy management and Bayesian deep learning.
\end{IEEEbiography}
\vspace{-2em}
\begin{IEEEbiography}[{\includegraphics[width=1in,height=1.25in,clip,keepaspectratio]{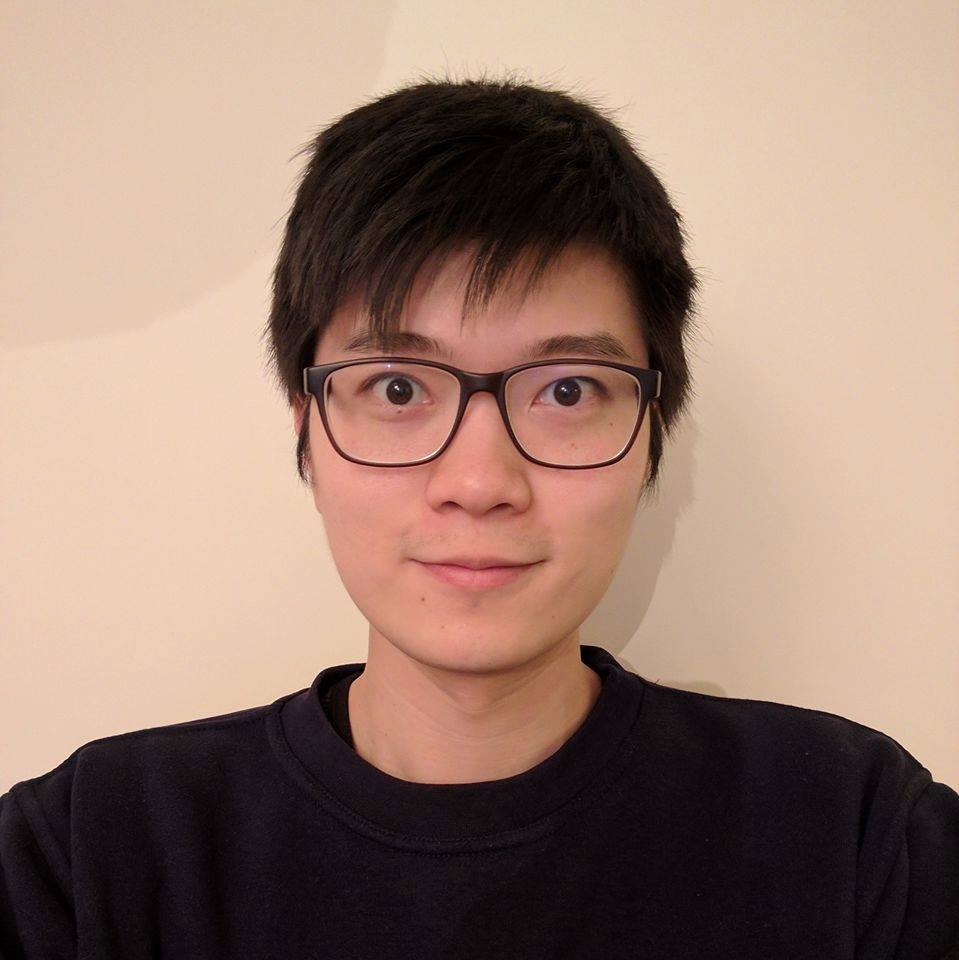}}]{Pei-Hao (Eddy) Su}
%Pei-Hao (Eddy) Su 
 is a PhD candidate under the supervision of Professor Steve Young in Dialogue Systems Group at Cambridge University. His research interests centre on applying deep learning, reinforcement learning and Bayesian approaches to dialogue management and reward estimation, with the aim of building systems that can learn directly from human interaction.
He has published around 30 peer-reviewed papers across top speech and NLP conferences and he received the best student paper award at ACL 2016.
\end{IEEEbiography}
\vspace{-2em}
\begin{IEEEbiography}[{\includegraphics[width=1in,height=1.25in,clip,keepaspectratio]{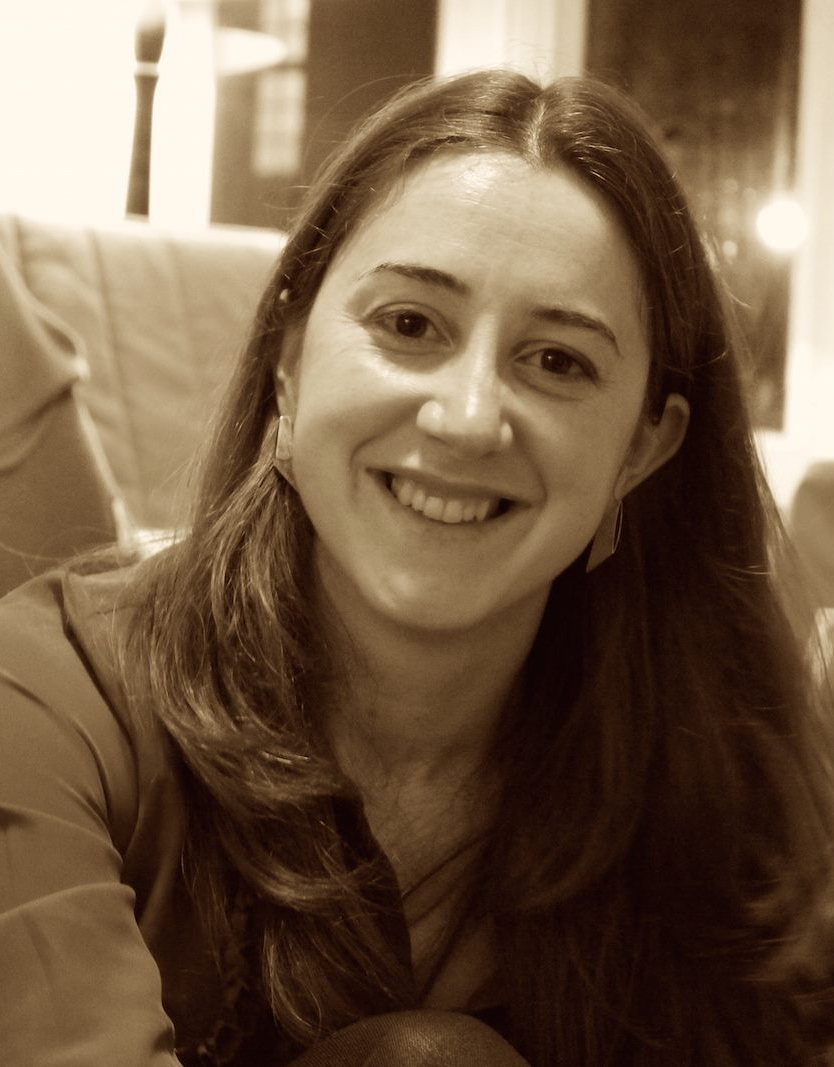}}]{Milica Ga\v{s}i\'{c}}
%Milica Ga\v{s}i\'{c}
is a Lecturer in Spoken Dialogue Systems at the Cambridge University Engineering Department and a Fellow of Murray Edwards College.  She  holds a BS in  Computer  Science and  Mathematics  from  the  University  of  Belgrade (2006), an  MPhil in Computer  Speech,  Text  and  Internet  Technology (2007) and a PhD in Engineering from the University of Cambridge (2011). The topic of her PhD was statistical dialogue modelling and she was awarded an EPSRC PhD plus award for her dissertation.  She has published around 50 journal articles and peer reviewed conference papers and received a number of best paper awards.  She is an elected committee member of IEEE SLTC and an appointed board member of Sigdial.
\end{IEEEbiography}

\end{document}